\documentclass[preprints,article,accept,moreauthors,pdftex]{mdpi} 

\firstpage{1} 
\pubvolume{1}
\issuenum{1}
\articlenumber{1}
\pubyear{2021}
\copyrightyear{2021}
\externaleditor{Academic Editor: Giovanni Dimauro} 
\datereceived{21 February 2021} 
\dateaccepted{19 March 2021} 
\datepublished{} 
\hreflink{https://doi.org/} 

\Title{Aerial Images Processing for Car Detection using Convolutional Neural Networks: Comparison between Faster R-CNN and YoloV3}

\TitleCitation{Aerial Images Processing for Car Detection using Convolutional Neural Networks: Comparison between Faster R-CNN and YoloV3}

\Author{Adel Ammar 
 $^{1,}$*\orcidA{}, Anis Koubaa $^{1,2,}$*\orcidB{}, Mohanned Ahmed $^{1}$\orcidC, Abdulrahman Saad $^{1}$\orcidD{} and {Bilel Benjdira $^{1,3}$}}

\AuthorNames{Adel Ammar, Anis Koubaa, Mohanned Ahmed and Abdulrahman Saad}
\AuthorCitation{Ammar, A.; Koubaa, A.; Ahmed, M.; Saad, A.; Benjdira, B.}
\address{%
$^{1}$ \quad Department of Computer Science, College of Computer \& Information Sciences, Prince Sultan University, 11586 Riyadh, 
 Saudi Arabia; mohanned.ahmed@riotu-lab.org (M.A.); abdelrahman.saad@riotu-lab.org (A.S.); bbenjdira@psu.edu.sa (B.B.) 
\\
$^{2}$ \quad CISTER Research Centre, ISEP, Polytechnic Institute of Porto, 4200-465 Porto, Portugal \\ 
$^{3}$ \quad SEICT Lab, LR18ES44, Enicarthage, University of Carthage, Tunis 1054, Tunisia} 

\corres{Correspondence: aammar@psu.edu.sa (A.A.); akoubaa@psu.edu.sa (A.K.)}

\abstract{This paper addresses the problem of car detection from aerial images using Convolutional Neural Networks (CNNs). This problem presents additional challenges as compared to car (or any object) detection from ground images because the features of vehicles from aerial images are more difficult to discern. To investigate this issue, we assess the performance of three state-of-the-art CNN algorithms, namely Faster R-CNN, which is the most popular region-based algorithm, as well as YOLOv3  {and YOLOv4, which are }known to be the fastest detection algorithms.~We analyze two datasets with different characteristics to check the impact of various factors, such as the UAV's (unmanned aerial vehicle) 
 altitude, camera resolution, and object size. A total of {52} training experiments were conducted to account for the effect of different hyperparameter values. The objective of this work is to conduct the most robust and exhaustive comparison between these three cutting-edge algorithms on the specific domain of aerial images. By using a variety of metrics, we show that the difference between YOLOv4 and YOLOv3 on the two datasets is statistically insignificant in terms of {Average Precision (AP)  (contrary to what was obtained on the COCO dataset). However, both of them yield markedly better performance than Faster R-CNN in most configurations. The only exception is that both of them exhibit a lower recall when object sizes and scales in the testing dataset differ largely from those in the training dataset.}} 


\begin{document}

\section{Introduction} \label{sec1}

Unmanned aerial vehicles (UAVs) are nowadays a key enabling technology for a large number of applications such as surveillance~\cite{benjdira2019conf}, tracking~\cite{dronetrack}, disaster management~\cite{Alotaibi}, smart parking~\cite{MultiTaskCNN2019}, and {Intelligent Transportation Systems~\cite{UAV-ITS}}, to name a few. Thanks to their versatility, UAVs offer unique capabilities in collecting   visual data using high-resolution cameras from different locations, angles, and altitudes. These capabilities provide rich datasets of images that can be analyzed to extract useful information that serves the purpose of the underlying applications.~{Compared to ground images, }UAV aerial imagery collection presents several advantages, including a large field of view, high spatial resolution, flexibility, and high mobility. Although satellite imagery also provides a bird's eye view of the earth, UAV-based aerial imagery presents several advantages as compared to satellite imagery.~In fact, UAV imagery has a much lower cost and provides more updated views (many satellite maps are several months old and do not present recent changes). In addition, it can be used for real-time image/video stream analysis in a much more affordable means. Aerial images have different resolutions as compared to satellite images. For example, in our experiments, we reached a resolution of 2 cm/pixel (and can have even {lower}) for aerial images using typical DJ (Shenzhen DJI Sciences and Technologies Ltd. https://www.dji.com) 
 drones, whereas satellite images have resolutions of {about} 15 cm/pixel as for the dataset described in~\cite{mundhenk2016large} and can be even larger.  

With the current hype of artificial intelligence and deep learning, there has been an increasing trend since 2012 (the birth of AlexNet) to use Convolutional Neural Networks (CNNs) to extract information from images and video streams. While CNNs have been proven to be the best approach for classification, detection, and semantic segmentation of images, {aerial images have many peculiarities that differ from the classical types of images (ground-level images). For example, objects can be viewed from different altitudes and viewpoints. Hence, a single class can have many patterns and representations to be learned. This is defined as high intra-class variance and indicates   high variability in the appearances of objects belonging to the same class. Moreover, different classes can share comparable  appearances, especially in high altitudes. This is defined as low inter-class variance and makes the learning task more challenging.}

Recently, there have been several research works that address  the problem of car detection from aerial images~\cite{Liu2015, doi:10.1080/01431161.2020.1757782, Audebert_2017, 8708261}.~In our previous work~\cite{benjdira2019conf}, we compared   YOLOv3 and Faster R-CNN in detecting cars from aerial images. However, we only used one small dataset from low-altitude UAV images collected at the premises of Prince Sultan University. However, the altitude at which the image is taken plays an essential role in the accuracy of   detection. 
In addition, we did not profoundly analyze advanced and essential performance metrics such as Intersection over Union (IoU) and the Mean Average Precision (mAP). In this paper, we address the gap, we consider multiple datasets with different configurations, {and we also compare   the newly released YOLOv4 object detector}. Our objective is to present a more comprehensive analysis of the comparison between these three state-of-the-art approaches {(Faster R-CNN, YOLOv3, and YOLOv4)}. 

In~\cite{MultiTaskCNN2019},  the authors mentioned the challenges faced with aerial images for car detection, namely the problem of having small objects and complex backgrounds. They addressed the problem with the proposed Multi-task Cost-sensitive-Convolutional Neural Network based on Faster R-CNN. Other researchers have addressed the problem applying deep learning techniques on aerial images, in  such contexts   as object detection and classification~\cite{sevo2016, SALDANAOCHOA201953}, semantic segmentation~\cite{Kampffmeyer2016, Azimi2019, Mou2018}, and generative adversarial networks (GANs)~\cite{Benjdira2019Segmentation}.

{Jiao et al.~\cite{jiao2019survey}   surveyed a large number of object detectors and reported their results on the COCO dataset~\cite{COCO}. Our objective in this paper is   different, since we focused on the depth-wise aspect of the comparison by selecting three recent algorithms that are  representative of the two main categories of object detectors, namely Faster R-CNN~\cite{Faster_R-CNN_journal} (a {two}-stage detector) as well as YOLOv3~\cite{YOLOv3}  and YOLOv4~\cite{yolov4} ({one}-stage detectors), examining a wide range of hyperparameters and assessing the effect of the size and characteristics of aerial view datasets}. The contributions of this paper are as follows: First, we consider two different datasets of aerial images for the car detection problem with different characteristics to investigate the impact of dataset  properties on the performance of the algorithms. In addition, we provide a thorough comparison between the {three} most sophisticated categories of CNN approaches for object detection, Faster RCCN, which is a region-based approach proposed in 2017, YOLOv3, which is {still the most popular} version of the You-Look-Only-Once approach proposed by Joseph Redmon in 2018, {and the latest version YOLOv4, released by Bochkovskiy et al., in April 2020.} 

The remainder of this paper is organized as follows. Section~\ref{sec2} discusses the related works that deal with car detection and aerial image analysis using CNN, and some comparative studies applied to other object detections. Section~\ref{sec3} sets forth the theoretical background of the {three} algorithms. Section~\ref{sec4} describes the datasets and the obtained results. Finally, Section~\ref{sec5} draws the main conclusions of this study.

\section{Related Works} \label{sec2}

Various techniques have been proposed in the literature to solve the problem of car detection in aerial images and similar related issues. The main challenge is the small size and the large number of objects  in aerial views, which may lead to information loss when performing convolution operations, as well as a difficulty to discern features because of the angle of view. {There are specific challenges for each type of aerial imagery (fixed CCTV cameras, satellite, or UAV), due to their disparate level of resolution. We present here the most recent, relevant works in object detection for each of these three imagery types, and we then highlight the value added of the present work.}

{\subsection{Fixed Surveillance Cameras}}

Xi et al.~\cite{MultiTaskCNN2019} addressed the problem of vehicle detection from overhead surveillance images. They proposed a multi-task approach based on the Faster R-CNN algorithm to which they added a cost-sensitive loss. The main idea was to subdivide the object detection task into simpler subtasks with enlarged objects, thus improving the detection of small objects that are frequent in aerial views.~In addition, the cost-sensitive loss gives more importance to   objects that are difficult to detect or occluded because of a complex background and aims at improving the overall performance. Their method outperformed state-of-the-art techniques on their own specific, {private} dataset that was collected from surveillance cameras placed on top of buildings surrounding a parking lot. However, their approach has not been tested on other datasets, nor on UAV images.

In a similar application, Kim et al.~\cite{kim2018comparison} compared various implementations of {CNN-based object detectors, namely} YOLO {(see Section~\ref{section-yolo})}, the {Single Shot MultiBox Detector} (SSD),  the {region-based  convolutional  neural network} (R-CNN), the  region-based  Fully  Convolutional  Neural Network 
(R-FCN), and SqueezeDet {\cite{wu2017squeezedet} (based on a Fully Convolutional Neural Network).~They applied these algorithms} on the problem of person detection, and trained and tested them on their own in-house dataset composed of images that were captured by surveillance cameras in retail stores. They found that YOLOv3 (with a 416 input size) and SSD ({with a VGG-500 feature extractor})~\cite{SSD} provide the best tradeoff between accuracy and response~latency.

In~\cite{Hardjono2018}, Hardjono et al. investigated the problem of automatic vehicle counting in CCTV images collected from four datasets with various resolutions. {On the one hand, they tested two} classical image processing techniques: Background Subtraction {(which calculates a foreground mask by subtracting a background model from the image)} and  the Viola Jones Algorithm {\cite{viola2001rapid}  (combining Haar-like Features, Integral Images, AdaBoost Algorithm~\cite{freund1997decision}, and Cascading Classifier), with Median or Gaussian Filters}. {On the other hand, they also applied} deep learning neural networks, namely YOLOv2~\cite{YOLOv2} and FCRN Fully Convolutional Regression Network) 
~\cite{FCRN}. Their results show that deep learning techniques yield markedly better detection results (in terms of F1 score) when applied on higher resolution datasets.

{\subsection{Satellite Imagery}}

Chen et al.~\cite{Chen2014} applied a technique based on a 
Hybrid Deep Convolutional Neural Network
 (HDNN) and a sliding window search to solve the vehicle detection problem {from Google Earth images}. The maps of particular layers of the CNN {(last convolutional layer and max-pooling layer)} are split into blocks of variable field sizes, so as to be able to extract features of various scales. {In addition, they modified the sliding windows to contain the main part of the vehicle to be detected. Thus,} they obtained an improved detection rate compared to the traditional deep architectures at that time, but with the expense of a high execution time (7~s per image, using a GPU).

{For the aim of car counting, Mundhenk et al.~\cite{mundhenk2016large} built their own Cars Overhead with Context (COWC) dataset containing 32,716 unique cars and 58,247 negative targets, standardized to a resolution of 15 cm per pixel, and annotated using single pixel points. The authors used a Convolutional Neural Network 
 that they called ResCeption, based on Inception synthesized with Residual Learning. {The main modification to the Inception architecture is the substitution of 1 $\times$ 1 convolutions by residual projection shortcuts}. The model was able to count the number of cars in test patches with a root mean square error of 0.66 at 1.3 FPS (frames per second)}.\\ 

\subsection{UAV Imagery}
{Relatively fewer works have addressed the problem of car detection from UAV images. }Ammour et al.~\cite{Ammour2017} used a pre-trained CNN coupled with a linear support vector machine (SVM) classifier to detect and count cars in high-resolution UAV images of urban areas. First, the input image is segmented into candidate regions using the mean-shift algorithm. The  VGG16~\cite{vgg} CNN model is then applied to windows that are extracted around each candidate region to generate descriptive features{, that are subsequently} classified using a linear SVM binary model. {Finally, they applied a fine-tuning morphological dilation for smoothing the detected regions.} This {multi-stage} technique achieved state-of-the-art performance {on a reduced testing dataset (5 images containing 127 car instances),} but it still falls short of real-time processing, mainly due to the high computational cost of the mean-shift segmentation stage. 

Liu and Mattyus~\cite{Liu2015} focused on fine-grained car detection. They {used a soft-cascade structure of integral channel features~\cite{dollar2009integral} to classify} car orientations and types {(car or truck)} in a dataset of aerial images of the city of Munich {consisting of 20 images taken at an altitude of 1000~m with a resolution of 5616 $\times$ 3744 and a GSD (Ground Sampling Distance) 
 of 13~cm. They obtained} an accuracy of 98\% {at a processing time of 4.4 s per image, which is faster than traditional techniques such as Viola Jones, but still far from real time. Such} classification can be used for  the {urban  planning, traffic management,} census estimation, and sociological analysis of cities and countries.

{\subsection{Our Contribution}}

{Table~\ref{tab:comparison-related-works} summarizes the datasets, algorithms, and results of the most similar related works on car detection, compared to the present paper.} The closest work to the present study is that of Benjedira et al.~\cite{benjdira2019conf} who presented a performance evaluation of Faster R-CNN and YOLOv3 algorithms, on a reduced UAV imagery dataset of cars. The present paper is an improvement over this work from several aspects:
\vspace{3pt}

\begin{enumerate}
\item[(1)] We use two datasets with different characteristics for training and testing, whereas most previous works described above tested their technique on a single proprietary dataset. {We show that annotation errors in the dataset have an important effect on the detection performance.}

\item[(2)] We added a third algorithm (YOLOv4) to the comparative analysis.

\item[(3)] We tested various hyperparameter values (three different input sizes for YOLOv3 {and YOLOv4 each}, two different feature extractors for Faster R-CNN, and various values of score and IoU thresholds).

\item[(4)] We conducted a more detailed comparison of the results, by showing the AP at different values of IoU thresholds, comparing the tradeoff between AP and inference speed, and calculating several new metrics that have been suggested for the COCO dataset~\cite{COCO}.
\end{enumerate}

\begin{figure}[H]
\widefigure
\includegraphics[width=18cm]{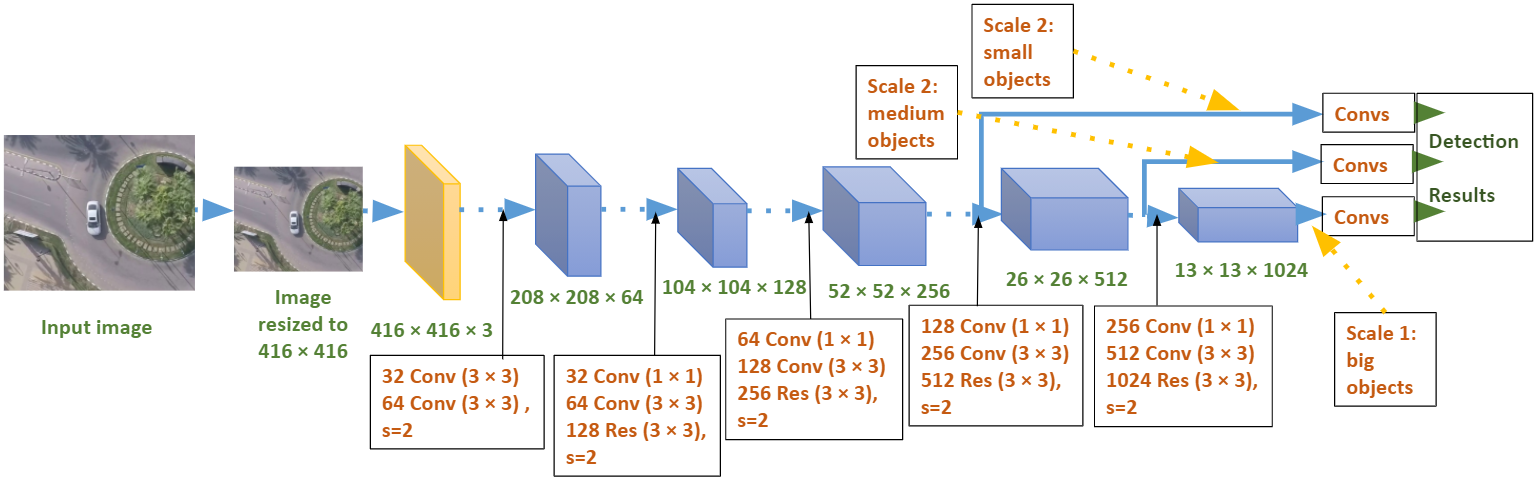}
\vspace{3pt}
\caption{YOLOv3 architecture.}
\label{fig:yolo-v3-architecture}  
\end{figure}

\begin{figure}[H]
\widefigure
\includegraphics[width=15cm]{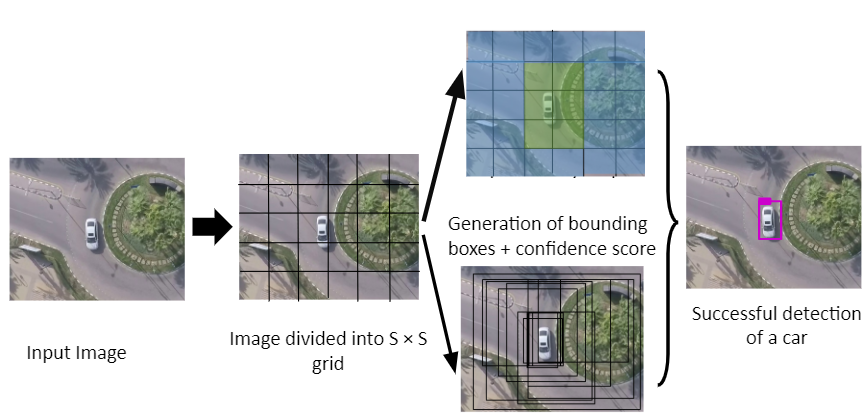}
\caption{Successive stages of the YOLOv3 model applied on car detection.
\label{fig:yolo-stages}}  
\end{figure}

\clearpage
\end{paracol}
\nointerlineskip
\begin{specialtable}[H]
  \tablesize{\fontsize{6.2pt}{6.2pt}\selectfont}
\widetable
\caption{{Comparison} of our paper with the related works.}
\label{tab:comparison-related-works} 
\setlength{\cellWidtha}{\columnwidth/4-2\tabcolsep-0.65in}
\setlength{\cellWidthb}{\columnwidth/4-2\tabcolsep-0.2in}
\setlength{\cellWidthc}{\columnwidth/4-2\tabcolsep-0.2in}
\setlength{\cellWidthd}{\columnwidth/4-2\tabcolsep-0.2in}
\scalebox{1}[1]{\begin{tabularx}{\columnwidth}{>{\PreserveBackslash\raggedright\arraybackslash}m{\cellWidtha}>{\PreserveBackslash\raggedright\arraybackslash}m{\cellWidthb}>{\PreserveBackslash\raggedright\arraybackslash}m{\cellWidthc}>{\PreserveBackslash\raggedright\arraybackslash}m{\cellWidthd}}
\toprule
\multicolumn{1}{l}{\textbf{Ref.}}                                               & \multicolumn{1}{l}{\textbf{Dataset Used}}                                                                                                                                                                                                                                                                                                                                                                                           & \multicolumn{1}{l}{\textbf{Algorithms}}                                                                                                                                                        & \multicolumn{1}{l}{\textbf{Main Results}}                                                                                                                                                                                                                                                                                                                                                      \\ \midrule
\textbf{\cite{mundhenk2016large} Mundhenk et al., 2016}                                     & \begin{tabular}[c]{@{}l@{}}Cars Overhead with Context (COWC):\\ 32,716 unique cars.\\ 58,247 negative targets.\\ 308,988 training patches and 79,447 testing patches.\\ Annotated using single pixel points.\\ Resolution: {1024 $\times$ 1024 and 2048 $\times$ 2048}.\end{tabular}                                                                                                                           & \multicolumn{1}{l}{\begin{tabular}[c]{@{}l@{}}ResCeption \\(Inception with Residual Learning)\end{tabular}}                                                                                                                                          & \begin{tabular}[c]{@{}l@{}}Up to 99.14\% correctly classified patches \\(containing cars or not).\\ F1 score of 94.34\% for detection.\\ Car counting: RMSE of 0.676.\end{tabular}                                                                                                                                                                                   \\ \midrule
\multicolumn{1}{l}{\textbf{\cite{MultiTaskCNN2019} Xi et al., 2019}}                          & \multicolumn{1}{l}{\begin{tabular}[c]{@{}l@{}}Parking lot dataset from aerial view.\\ Training: 2000 images.\\ Testing: 1000 images.\\ Number of instances: NA.\\ Resolution: 5456 $\times$ 3632.\end{tabular}}                                                                                                                                                                                                                              & \multicolumn{1}{l}{\begin{tabular}[c]{@{}l@{}}Multi-Task Cost-sensitive \\ Convolutional Neural Network \\(MTCS-CNN).\end{tabular}}                                                              & \multicolumn{1}{l}{mAP of 85.3\% for car detection.}                                                                                                                                                                                                                                                                                                                                           \\ \midrule
\multicolumn{1}{l}{\textbf{\cite{Chen2014} Chen et al., 2014}}                        & \multicolumn{1}{l}{\begin{tabular}[c]{@{}l@{}}63 satellite images collected from Google Earth.\\ Training: 31 images (3901 vehicles). \\ Testing: 32 images (2870 vehicles).\\ Resolution: 1368 $\times$ 972.\end{tabular}}                                                                                                                                                                                                                  & \multicolumn{1}{l}{\begin{tabular}[c]{@{}l@{}}Hybrid Deep Convolutional \\ Neural Network (HDNN).\end{tabular}}                                                                                & \multicolumn{1}{l}{Precision up to 98\% at a recall rate of 80\%.}                                                                                                                                                                                                                                                                                                                             \\ \midrule
\multicolumn{1}{l}{\textbf{\cite{Ammour2017} Ammour et al., 2017}}                      & \multicolumn{1}{l}{\begin{tabular}[c]{@{}l@{}}8 images acquired by UAV.\\ Training: 3 images (136 positive instances, \\and 1864 negative instances).\\ Testing: 5 images (127 positive instances). \\ Resolution: \\ Variable from 2424 $\times$ 3896 to 3456 $\times$ 5184.\\ Spatial resolution of 2 cm.\end{tabular}}                                                                                                                               & \multicolumn{1}{l}{\begin{tabular}[c]{@{}l@{}}Pre-trained CNN coupled with \\ a linear support vector machine (SVM).\end{tabular}}                                                             & \multicolumn{1}{l}{\begin{tabular}[c]{@{}l@{}}Precision from 67\% up to 100\%, \\and recall from 74\% up to 84\%, \\ on the five testing images.\\ Inference time: \\ between 11 and 30 min/image.\end{tabular}}                                                                                                                                                                                 \\ \midrule
\multicolumn{1}{l}{\textbf{\cite{Hardjono2018} Hardjono et al., 2018}}                    & \multicolumn{1}{l}{\begin{tabular}[c]{@{}l@{}}4 CCTV datasets:\\ - Dataset 1: 3 s videos at 1 FPS.\\ Resolution: 480 $\times$ 360\\ - Dataset 2: 60 min:32 sec video at 9 FPS.\\ Resolution: 1920 $\times$ 1080\\ - Dataset 3: 30~min:27~sec video at 30 FPS.\\ Resolution: 1280 $\times$ 720\\ - Dataset 4: 32~sec video at 30 FPS.\\ Resolution: 1280 $\times$ 720\\ Training: 1932 positive instances \\ and 10,000 negative instances.\end{tabular}} & \multicolumn{1}{l}{\begin{tabular}[c]{@{}l@{}}- Background Subtraction (BS)\\ - Viola Jones (VJ)\\- YOLOv2\end{tabular}}                                                                         & \multicolumn{1}{l}{\begin{tabular}[c]{@{}l@{}}- BS: \\ F1 score from 32\% to 55\%. \\ Inference time from 23 to 40 ms.\\ - VJ: \\ F1 score from 61\% to 75\%. \\ Inference time from 39 to 640 ms.\\ - YOLOv2: \\F1 score from 92\% to 100\% on Datasets 2 to 4. \\Inference time not reported.\end{tabular}}                                                                          \\ \midrule
\multicolumn{1}{l}{\textbf{\cite{benjdira2019conf} Benjdira et al., 2019}}                & \multicolumn{1}{l}{\begin{tabular}[c]{@{}l@{}}PSU+{[}27{]} UAV dataset:\\ Training: 218 images (3365 car instances).\\ \\ Testing: 52 images (737 car instances).\\ \\ Resolution: \\ Variable from 684 $\times$ 547 to 4000 $\times$ 2250.\end{tabular}}                                                                                                                                                                                            & \multicolumn{1}{l}{\begin{tabular}[c]{@{}l@{}}- YOLOv3 (input size: 608 $\times$ 608).\\ \\ - Faster R-CNN \\(Feature extractor: Inception ResNet v2).\end{tabular}}                                & \multicolumn{1}{l}{\begin{tabular}[c]{@{}l@{}}- YOLOv3: \\ F1 score of 99.9\%. \\ Inference time: 57 ms.\\ - Faster R-CNN: \\ F1 score of 88\%. \\ Inference time: 1.39 s.\\ (Using an Nvidia GTX 1080 GPU).\end{tabular}}                                                                                                                                                                     \\ \midrule 
\multicolumn{1}{l}{\textbf{\begin{tabular}[c]{@{}l@{}}Our paper\end{tabular}}} & \multicolumn{1}{l}{\begin{tabular}[c]{@{}l@{}}- Stanford UAV dataset:\\ Training: 6872 images (74,826 car instances).\\ Testing: 1634 images (8131 car instances).\\ Resolution: Variable from 1184 $\times$ 1759 to 1434 $\times$ 1982.\\ \\PSU+{[}27{]} UAV dataset: \\ Training: 218 images (3365 car instances).\\ Testing: 52 images (737 car instances).\\ Resolution: Variable from 684 $\times$ 547 to 4000 $\times$ 2250.\end{tabular}}                     & \multicolumn{1}{l}{\begin{tabular}[c]{@{}l@{}}- YOLOv3 {and YOLOv4}\\(input sizes: 320 $\times$ 320, 416 $\times$ 416, and 608 $\times$ 608).\\ \\ - Faster R-CNN \\ (Feature extractors: Inception v2, and Resnet50).\end{tabular}} & \multicolumn{1}{l}{\begin{tabular}[c]{@{}l@{}}{- YOLOv4:} \\ {F1 score: up to 34.4\% on the Stanford dataset}\\                 {up to 94.6\% on the PSU dataset.} \\{ Inference time: from 45 to 80 ms.}\\- YOLOv3: \\ F1 score: up to 32.6\% on the Stanford dataset\\                 up to 96.0\% on the PSU dataset. \\ Inference time: from 43 to 85 ms.\\ - Faster R-CNN: \\ F1 score: up to 31.4\% on the Stanford dataset\\                 up to 84.5\% on the PSU dataset. \\ Inference time: from 52 to 160 ms.\\ (Using an Nvidia GTX 1080 GPU).\end{tabular}} \\ \bottomrule
\end{tabularx}}
\end{specialtable}
\begin{paracol}{2}
\switchcolumn

\vspace{-12pt}

\section{Theoretical Overview of Faster R-CNN and YOLO Architectures} \label{sec3}

Object detection is an old fundamental problem in image processing, for which various approaches have been applied. However, since 2012, deep learning techniques have markedly outperformed classical ones. {The object detection algorithms based on deep learning are classified into two large branches: two-stage detectors and one-stage detectors. From each   of these two branches, we selected, in this study, the best performing algorithms. We selected in the first branch, Faster R-CNN~\cite{Faster_R-CNN_journal}{, which is the most representative model from the two-stage family, according to~\cite{carranza2021performance}}. In the second branch, we selected  the YOLO algorithm and picked out its most recent versions: YOLO v3~\cite{YOLOv3} and YOLO v4~\cite{yolov4}. The selected algorithms have been proven successful in terms of of accuracy and speed in a wide variety of applications. }.


\subsection{Two-Stage Detector: Faster R-CNN} \label{section-frcnn}


R-CNN, as coined by~\cite{R-CNN}, is a Convolutional Neural Network (CNN) combined with a region-proposal algorithm that hypothesizes object locations. It initially extracts a fixed number of regions (2000), by means of a selective search. It then merges similar regions together, using a greedy algorithm, to obtain the candidate regions on which the object detection will be applied. Afterwards, the same authors proposed an enhanced algorithm called Fast R-CNN~\cite{Fast_R-CNN} by using a shared convolutional feature map that the CNN generates directly from the input image, and from which the regions of interest (RoI) are {extracted}. Finally, Ren et al.~\cite{Faster_R-CNN_journal} proposed a Faster R-CNN algorithm that introduced a Region Proposal Network (RPN), which is a dedicated fully convolutional neural network that is trained end-to-end (Figure~\ref{fig:RPN_training}) to predict both object bounding boxes and objectness scores in an almost computationally cost-free manner (around 10 ms per image). This important algorithmic change thus replaced the selective search algorithm, which was very computationally expensive and represented a bottleneck for previous object detection deep learning systems.  As a further optimization, the RPN ultimately shares the convolutional features with the Fast R-CNN detector, after   first being independently trained. 
For training the RPN, Faster R-CNN kept the multi-task loss function already used in Fast R-CNN~\cite{Fast_R-CNN}.

Faster R-CNN uses three scales and three aspect ratios for every sliding position, and is translation-invariant. In addition, it conserves the aspect ratio of the original image while resizing it, so that one of its dimensions is 1024 or 600.


\begin{figure}[H]
\includegraphics[width=8cm, height=10cm]{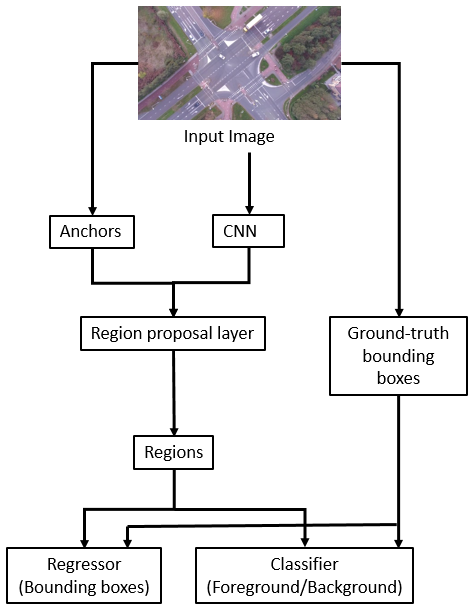}
\caption{Region Proposal Network (RPN) architecture.
\label{fig:RPN_training}}  
\end{figure}

{\subsection{One-Stage Detectors}}
{We have considered two networks of the one-stage category: YOLOv3  and YOLOv4. We first describe the architecture of YOLOv3 and then briefly enumerate the enhancements made in YOLOv4.}\\

\subsubsection{YOLOv3} \label{section-yolo}
Contrary to R-CNN variants, YOLO~\cite{YOLO2016}, which is an acronym for You Only Look Once, does not extract region proposals, but processes the complete input image only once using a Fully Convolutional Neural Network
 that predicts the bounding boxes and their corresponding class probabilities, based on the global context of the image.~The first version was published in 2016.~Later on in 2017, a second version, YOLOv2~\cite{YOLOv2}, was proposed, which introduced batch normalization, a retuning phase for the classifier network, and dimension clusters as anchor boxes for predicting bounding boxes. Finally, in 2018, YOLOv3~\cite{YOLOv3} improved the detection further by adopting several new features:
\begin{itemize}
    \item Replacing the mean squared error by cross-entropy for the loss function. The cross-entropy loss function is calculated as follows: 
    \begin{equation}\label{eq:2}
    -\sum_{c=1}^M \delta_{x\in c} log(p(x\in c))
    \end{equation}
    where M is the number of classes, c is the class index, x is an observation, $\delta_{x\in c}$ is an indicator function that equals 1 when $c$ is the correct class for the observation $x$, and $log(p(x\in c))$ is the natural logarithm of the predicted probability that observation $x$ belongs to class $c$. 
    \item Using logistic regression (instead of the softmax function) for predicting an objectness score for every bounding box. 
    \item Using a significantly larger feature extractor network with 53 convolutional layers (Darknet-53 replacing Darknet-19). It consists mainly of 3 $\times$ 3 and 1 $\times$ 1 filters, with some skip connections {(Figure~\ref{fig:yolo-v3-architecture}}) inspired from ResNet~\cite{ResNet}.
\end{itemize}

Contrary to Faster R-CNN's approach, each ground-truth object in YOLOv3 is assigned only one bounding box prior. These successive variants of YOLO were developed with the objective of obtaining a maximum mAP while keeping the fastest execution that makes it suitable for real-time applications. Special emphasis has been put on execution time, so that YOLOv3 is equivalent to state-of-the-art detection algorithms such as SSD~\cite{SSD} in terms of accuracy but with the advantage of being three times faster~\cite{YOLOv3}. Figure~\ref{fig:yolo-stages} depicts the main stages of  the YOLOv3 algorithm when applied to the car detection problem. Variable input sizes are allowed in YOLO. We have tested the three input sizes {that are usually used (as in the original YOLOv3 paper~\cite{YOLOv3})}: 320 $\times$ 320, 416 $\times$ 416, and 608 $\times$ 608. 


\end{paracol}
\nointerlineskip


\begin{figure}[H]
\widefigure
\includegraphics[width=14cm]{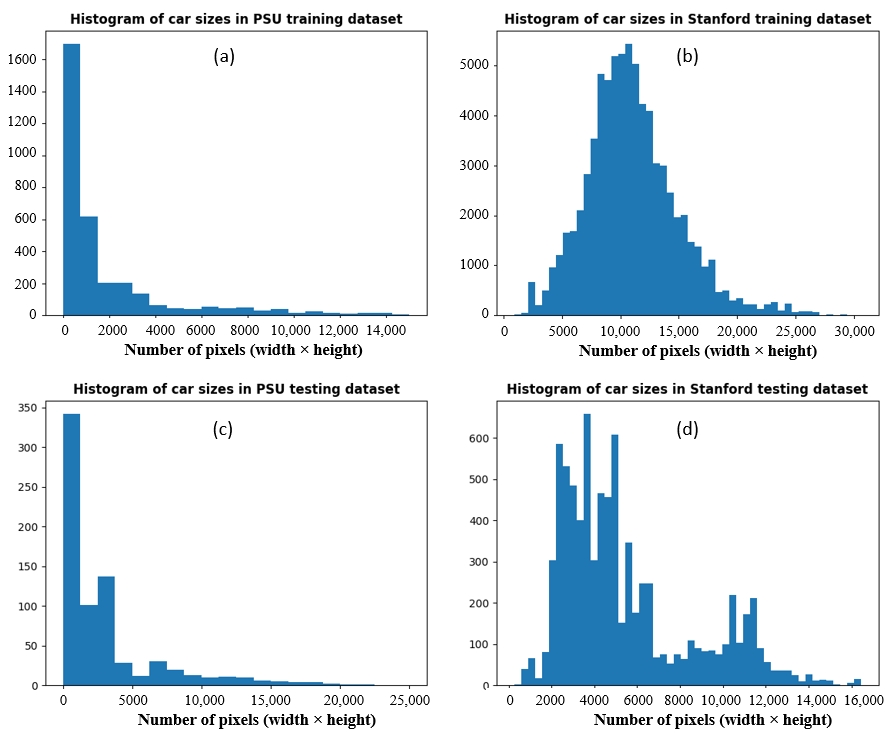}
\caption{{Histogram of car sizes in PSU 
 (a,c) and Stanford (b,d) training (a,b) and testing (c,d) datasets, expressed as the number pixels inside the ground truth bounding boxes (width $\times$ height)}. 
\label{fig:hist_car_sizes}}  

\end{figure}

\begin{paracol}{2}
\switchcolumn
\vspace{-6pt}

\subsubsection{YOLOv4} \label{section-yolo-v4}

YOLOv4~\cite{yolov4}  was introduced after two years of cumulative improvements over YOLOv3~\cite{YOLOv3}, leveraging the recent advances in deep learning. It achieves an accuracy of 43.5\% AP on the MS COCO dataset compared to 33.0\% AP for YOLOv3. This high accuracy is made while keeping a very efficient inference time (65 FPS on Tesla V100). YOLOv4 aims to make object detection run efficiently and smoothly on the low-cost hardware provided on most edge devices. 

Concerning the technical improvements made in YOLOv4, they are classified into two categories. The first category is named  the Bag of Freebies (BoF) and designates   improvements that can be made during   training without affecting the inference time. This includes  CutMix~\cite{CutMix} and Mosaic data augmentation techniques, DropBlock regularization~\cite{dropblock}, class label smoothing, Complete IoU (CIoU) loss~\cite{distanceiou}, Cross mini-Batch Normalization (CmBN)~\cite{CBN}, Self Adversarial Training (SAT), multiple anchors for a single ground truth, {cosine annealing scheduler}~\cite{loshchilov2016sgdr}, and optimal hyper-parameters obtained through genetic algorithms.

On the other hand, the second category is named Bag of Specials (BoS) and represents   improvements that slightly affect the inference time while making a considerable increase in accuracy. This includes the mish activation function~\cite{misra2019mish}, Cross Stage Partial connections (CSP))~\cite{cspnet}, Multi-input Weighted Residual Connection (MiWRC)~\cite{efficientdet},  the Spatial Pyramid Pooling (SPP) block~\cite{DC-SPP-YOLO},  the Spatial Attention Module (SAM) block~\cite{CBAM},  the Path Aggregation Network (PAN) block~\cite{PAN}, and  the Distance IoU Loss (DIoU)~\cite{distanceiou} used as a factor in the Non-Maximum-Suppression (NMS) step.

To summarize, Table~\ref{tab:comparison-YOLO-FRCNN} compares the features and parameters of Faster R-CNN, YOLOv3{, and YOLOv4}. While successive optimizations and mutual inspirations made the methodology of the two {architectures} relatively close, the main difference remains that Faster R-CNN has two separate phases of region proposals and classification (although now with shared features), whereas YOLO has always combined the classification and bounding-box regression processes.

\clearpage
\end{paracol}
\nointerlineskip
\begin{specialtable}[H]
\widetable
  \tablesize{\small}
\caption{{Theoretical comparison of YOLOv3, YOLOv4, and Faster R-CNN.}}
\label{tab:comparison-YOLO-FRCNN} 
\setlength{\cellWidtha}{\columnwidth/4-2\tabcolsep-0.7in}
\setlength{\cellWidthb}{\columnwidth/4-2\tabcolsep+0.3in}
\setlength{\cellWidthc}{\columnwidth/4-2\tabcolsep+0.2in}
\setlength{\cellWidthd}{\columnwidth/4-2\tabcolsep+0.2in}
\scalebox{1}[1]{\begin{tabularx}{\columnwidth}{>{\PreserveBackslash\raggedright\arraybackslash}m{\cellWidtha}>{\PreserveBackslash\raggedright\arraybackslash}m{\cellWidthb}>{\PreserveBackslash\raggedright\arraybackslash}m{\cellWidthc}>{\PreserveBackslash\raggedright\arraybackslash}m{\cellWidthd}}
\toprule

\textbf{} & \textbf{YOLOv3} & \textbf{YOLOv4} & \textbf{Faster R-CNN} \\ \midrule
\textbf{Phases} & \begin{tabular}[c]{@{}l@{}}Concurrent\\ bounding box regression,\\ and classification\end{tabular} & \begin{tabular}[c]{@{}l@{}}Concurrent\\ bounding box regression,\\ and classification\end{tabular} & \begin{tabular}[c]{@{}l@{}}RPN +\\ Fast R-CNN\\ object detector\end{tabular} \\ \midrule
\textbf{\begin{tabular}[c]{@{}l@{}}Neural \\ network \\ type\end{tabular}} & Fully convolutional. & Fully convolutional. & \begin{tabular}[c]{@{}l@{}}Fully convolutional\\ (RPN and 4\\ detection network).\end{tabular} \\ \midrule
\textbf{\begin{tabular}[c]{@{}l@{}}Backbone \\ feature \\ extractor\end{tabular}} & \begin{tabular}[c]{@{}l@{}}Darknet-53\\ (53 convolutional layers).\end{tabular} & \begin{tabular}[c]{@{}l@{}}CSPDarknet53\\ (53 convolutional layers).\end{tabular} & \begin{tabular}[c]{@{}l@{}}VGG-16 or\\ Zeiler \& Fergus(ZF).\\ Other feature extractors\\ can also be incorporated.\end{tabular} \\ \midrule
\textbf{\begin{tabular}[c]{@{}l@{}}Location\\ detection\end{tabular}} & \begin{tabular}[c]{@{}l@{}}Anchor-based\\ (dimension clusters).\end{tabular} & Anchor-based & Anchor-based \\ \midrule
\textbf{\begin{tabular}[c]{@{}l@{}}Number of \\ anchors \\ boxes\end{tabular}} & \begin{tabular}[c]{@{}l@{}}Only one bounding-box \\ prior for each \\ ground-truth object.\end{tabular} & \begin{tabular}[c]{@{}l@{}}Using multiple anchors\\ for a single ground truth\end{tabular} & \begin{tabular}[c]{@{}l@{}}3 scales and 3 aspect ratios,\\ yielding k = 9 anchors \\ at each sliding position.\end{tabular} \\ \midrule
\textbf{\begin{tabular}[c]{@{}l@{}}Default \\ Anchors\\ sizes\end{tabular}} & \begin{tabular}[c]{@{}l@{}}(10,13), (16,30), (33,23),\\ (30,61), (62,45), (59,119),\\ (116,90), (156,198), (373,326)\end{tabular} & \begin{tabular}[c]{@{}l@{}}(12,16), (19,36), (40,28),\\(36,75), (76,55), (72,146),\\(142,110), (192,243), (459,401)\end{tabular} & \begin{tabular}[c]{@{}l@{}}Scales: (128,128), (256,256),\\ (512,512). \\ Aspect ratios: 1:1, 1:2, 2:1.\end{tabular} \\ \midrule
\textbf{\begin{tabular}[c]{@{}l@{}}IoU \\ thresholds\end{tabular}} & One (at 0.5). & One (at 0.213) & Two (at 0.3 and 0.7). \\ \midrule
\textbf{\begin{tabular}[c]{@{}l@{}}Loss \\ function\end{tabular}} & Binary cross-entropy loss & Complete IoU loss: CIoU & \begin{tabular}[c]{@{}l@{}}Multi-task loss:\\ - Log loss for classification.\\ - Smooth L1 for regression.\end{tabular} \\ \midrule
\textbf{Input size} & \begin{tabular}[c]{@{}l@{}}Different possible input sizes\\ (n $\times$ n with n multiple of 32).\end{tabular} & \begin{tabular}[c]{@{}l@{}}Different possible input sizes\\ (n $\times$ n with n multiple of 32).\end{tabular} & \begin{tabular}[c]{@{}l@{}}- Conserves the aspect \\ ratio of the original image.\\ - Either the smallest\\ dimension is 600,\\ or the largest dimension\\ is 1024.\end{tabular} \\ \midrule
\textbf{Momentum} & Default value: 0.9. & Default value: 0.949 & Default value: 0.9. \\ \midrule
\textbf{\begin{tabular}[c]{@{}l@{}}Weight \\ decay\end{tabular}} & Default value: 0.0005. & Default value: 0.0005 & Default value: 0.0005. \\ \midrule
\textbf{\begin{tabular}[c]{@{}l@{}}Batch \\ size\end{tabular}} & Default value: 64. & Default value: 64. & Default value: 1. \\ \bottomrule
\end{tabularx}
}
\end{specialtable}
\begin{paracol}{2}
\switchcolumn
\vspace{-9pt}

\section{Experimental Comparison between Faster R-CNN {, YOLOv3, and YOLOv4}} \label{sec4}

{In this section, we will first describe the two datasets used for training and testing, and the hyperparameters chosen for each algorithm, and then present and discuss the results obtained.}

\subsection{Datasets} \label{section-datasets}
In order to obtain a robust comparison, we tested  the Faster R-CNN, YOLOv3{, and YOLOv4} algorithms on two datasets of aerial images showing completely different characteristics.
\begin{itemize}
    \item 
The Stanford dataset~\cite{robicquet2016learning} 
consists of a large-scale collection of aerial images and videos of a university campus containing various agents (cars, buses, bicycles, golf carts, skateboarders, and pedestrians). It was obtained using a 3DR 
 SOLO 
 quadcopter (equipped with a 4k camera) that flew over various crowded campus scenes, at an altitude of around 80 m. It is originally composed of eight scenes, but since we are exclusively interested in car detection, we chose only three scenes that contains the largest percentage of cars: Nexus {(in which 29.51\% of objects are cars), Gates (1.08\%), and DeathCircle (4.71\%).} All other scenes contain less than 1\% of cars. {We used the two first scenes for training  and the third one for testing.} In addition, we   removed images that contain no cars. {Table~\ref{tab:Stanford_and_PSU_Datasets} shows the number of images and instances in the training and testing datasets. The images in the selected scene have variable sizes, as shown in Table~\ref{tab:ImageSizeStanford}, and contain cars of various sizes, as depicted in Figure~\ref{fig:hist_car_sizes}. The average car size (calculated based on the ground-truth bounding boxes) is shown in Table~\ref{tab:average_car_size}. The discrepancy observed between the training and testing datasets in terms of car sizes is explained by the fact that we used different scenes for the training and testing datasets, as explained above. This discrepancy will constitute an additional challenge for the considered object detection algorithms. Furthermore, we noticed that the ground-truth bounding boxes in some images contain some mistakes (bounding boxes containing no objects) and imprecisions (many bounding boxes are much larger than the objects inside them), as can be seen in Figure~\ref{fig:Stanford-sample}, but we used them as they are in order to assess the impact of annotation errors on detection performance. In fact, the Stanford Drone Dataset was not primarily designed for object detection, but for  trajectory forecasting and tracking.} 
\end{itemize}
 
\end{paracol}
\nointerlineskip
\begin{specialtable}[H]
\widetable
\caption{Number of images and car instances in Stanford and PSU (Prince Sultan University) datasets.} 
\label{tab:Stanford_and_PSU_Datasets} 
\begin{tabular*}{\hsize}{@{}@{\extracolsep{\fill}}lcccccc@{}}
\toprule
                                              & \multicolumn{3}{c}{\textbf{Stanford Dataset}} & \multicolumn{3}{c}{\textbf{PSU Dataset}}   \\ \cmidrule{2-7} 
\multicolumn{1}{c}{}                         & \textbf{Training Set}  & \textbf{Testing Set}  & \textbf{Total}  & \textbf{Training Set} & \textbf{Testing Set} & \textbf{Total} \\ \midrule
\multicolumn{1}{l}{Number of images}        & 6872          & 1634         & 8506   & 218          & 52          & 270   \\ 
\multicolumn{1}{l}{Percentage}              & 80.8\%        & 19.2\%       & 100\%  & 80.7\%       & 19.3\%      & 100\% \\ 
\multicolumn{1}{l}{Number of car instances} & 74,826        & 8131        & 82,957 & 3364        & 738         & 4102 \\ \bottomrule
\end{tabular*}
\end{specialtable}
\begin{paracol}{2}
\switchcolumn

\vspace{-9pt}


\begin{specialtable}[H]
\caption{Image size in the Stanford dataset.}
\label{tab:ImageSizeStanford} 
\setlength{\cellWidtha}{\columnwidth/2-2\tabcolsep+0.0in}
\setlength{\cellWidthb}{\columnwidth/2-2\tabcolsep+0.0in}
\scalebox{1}[1]{\begin{tabularx}{\columnwidth}{>{\PreserveBackslash\centering}m{\cellWidtha}>{\PreserveBackslash\centering}m{\cellWidthb}}
\toprule
\multicolumn{1}{c}{\textbf{Size}} & \multicolumn{1}{c}{\textbf{Number of Images}} \\ \midrule
1409 $\times$ 1916                  & 1634                                  \\ 
1331 $\times$ 1962                  & 1558                                  \\ 
1330 $\times$ 1947                  & 1557                                  \\ 
1411 $\times$ 1980                  & 1494                                  \\ 
1311 $\times$ 1980                  & 1490                                  \\ 
1334 $\times$ 1982                  & 295                                   \\ 
1434 $\times$ 1982                  & 142                                   \\ 
1284 $\times$ 1759                  & 138                                   \\
1425 $\times$ 1973                  & 128                                   \\ 
1184 $\times$ 1759                  & 70                                    \\ \bottomrule
\end{tabularx}}
\end{specialtable}
\vspace{-9pt}

\begin{specialtable}[H]
\caption{{Average car width and length (in pixels) in the PSU 
 and Stanford datasets, calculated based on the ground-truth bounding boxes.}}
\label{tab:average_car_size} 
\setlength{\cellWidtha}{\columnwidth/3-2\tabcolsep+0.0in}
\setlength{\cellWidthb}{\columnwidth/3-2\tabcolsep+0.0in}
\setlength{\cellWidthc}{\columnwidth/3-2\tabcolsep+0.0in}
\scalebox{1}[1]{\begin{tabularx}{\columnwidth}{>{\PreserveBackslash\centering}m{\cellWidtha}>{\PreserveBackslash\centering}m{\cellWidthb}>{\PreserveBackslash\centering}m{\cellWidthc}}
\toprule
\textbf{Dataset}           & \multicolumn{1}{c}{\begin{tabular}[c]{@{}l@{}}\textbf{Average  Car Width}\end{tabular}} & \multicolumn{1}{c}{\begin{tabular}[c]{@{}l@{}}\textbf{Average Car Length}\end{tabular}} \\ \midrule
PSU training      & 48                                                                                & 36                                                                                 \\ 
PSU testing       & 55                                                                                & 46                                                                                 \\ 
Stanford training & 72                                                                                & 152                                                                                \\ 
Stanford testing  & 60                                                                                & 90                                                                                 \\ \bottomrule
\end{tabularx}}
\end{specialtable}

\clearpage
\end{paracol}
\nointerlineskip
\begin{paracol}{2}
\switchcolumn
\vspace{-6pt}

\begin{figure}[H]
\includegraphics[width=7.5cm]{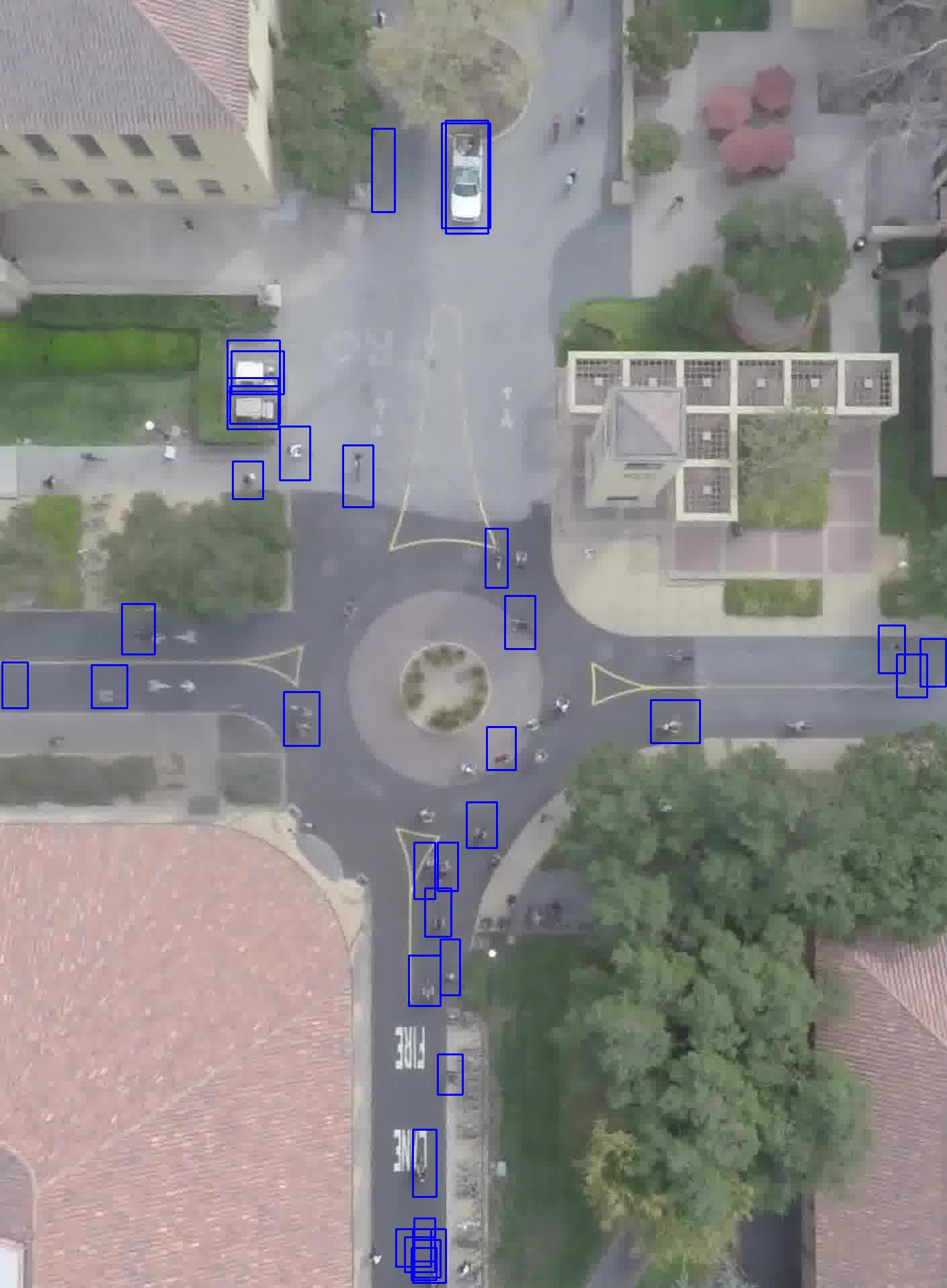}
\caption{A sample image of the Stanford dataset, with ground-truth bounding boxes showing some annotation errors and imprecisions.  
\label{fig:Stanford-sample}}  
\end{figure}

\begin{itemize}
    \item
The PSU dataset
was collected from two sources: an open dataset of aerial images available on Github~\cite{aerial-car-dataset} and our own images acquired after flying a 3DR SOLO drone equipped with a GoPro Hero 4 camera, in an outdoor environment at a PSU parking lot. The drone recorded videos from which frames were extracted and manually labeled. Since we are only interested in a single class, images with no cars were removed from the dataset. {The training/testing split was made randomly.}

Table~\ref{tab:Stanford_and_PSU_Datasets} shows the number of images and instances in the training and testing datasets. The dataset thus obtained contains images of different sizes, as shown in Table~\ref{tab:ImageSizePSU}{, and contains cars of various sizes, as depicted in Figure~\ref{fig:hist_car_sizes}. The average car size (calculated based on the ground-truth bounding boxes) in the training and testing datasets is shown in Table~\ref{tab:average_car_size}.} We have made this dataset available on~\cite{psu-dataset}.

\end{itemize}

\vspace{-3pt}

\begin{specialtable}[H]
\caption{Image size in the PSU dataset.} 
\label{tab:ImageSizePSU} 
\setlength{\cellWidtha}{\columnwidth/2-2\tabcolsep+0.0in}
\setlength{\cellWidthb}{\columnwidth/2-2\tabcolsep+0.0in}
\scalebox{1}[1]{\begin{tabularx}{\columnwidth}{>{\PreserveBackslash\centering}m{\cellWidtha}>{\PreserveBackslash\centering}m{\cellWidthb}}
\toprule
\multicolumn{1}{c}{\textbf{Size}} & \multicolumn{1}{c}{\textbf{Number of Images}} \\ \midrule
1920 $\times$ 1080 & 172                                   \\ 
1764 $\times$ 430  & 26                                    \\ 
684 $\times$ 547   & 21                                    \\ 
1284 $\times$ 377  & 20                                    \\ 
1280 $\times$ 720  & 19                                    \\ 
4000 $\times$ 2250 & 12                                    \\ \bottomrule
\end{tabularx}}
\end{specialtable}

\subsection{Hyperparameters} \label{hyperparams}
The main hyperparameter for YOLOv3 {and YOLOv4 networks} is the input size, for which we tested three values (320 $\times$ 320, 416 $\times$ 416, and 608 $\times$ 608){, as explained in \mbox{Section~\ref{section-yolo}}}. On the other hand, the main hyperparameter for Faster R-CNN is the feature extractor. We tested two different feature extractors: Inception-v2~\cite{BN} (also called BN-inception in the literature~\cite{bianco2018benchmark}) and Resnet50~\cite{ResNet}. {As explained in Section~\ref{section-frcnn}, the default setting of Faster R-CNN conserves the aspect ratio of the original image while resizing it, so that one of its dimensions is 1024 or 600. However, to be able to fairly compare its precision and speed with YOLO {algorithms}, which use fixed input sizes, we also tested Faster R-CNN with a fixed input size of 608 $\times$ 608, for each of the two feature extractors.} These settings make a total of {10} classifiers that we trained and tested on the two datasets described above, which amounts to {20} experiments,   summarized in Table~\ref{tab:my-table-experiments}. In these experiments, we kept the default values for the momentum (0.9), weight decay (0.0005), learning rate (initial rate of 10\textsuperscript{$-$3} for YOLOv3  {and YOLOv4}, 2 $\times$ 10\textsuperscript{$-$4} for Faster R-CNN with Inception-v2, and 3 $\times$ 10\textsuperscript{$-$4} with Resnet50), batch size (64 for YOLOv3 {and YOLOv4}, and 1 for Faster R-CNN), {and anchor sizes (see Table~\ref{tab:comparison-YOLO-FRCNN}).} {Furthermore, we conducted additional experiments with different values of learning rates (10\textsuperscript{$-$5}, 10\textsuperscript{$-$4}, 10\textsuperscript{$-$3}, and 10\textsuperscript{$-$2}) for each of the main algorithms (Faster R-CNN with Inception-v2, Faster R-CNN with Resnet 50, YOLOv3, and {YOLOv4} with  the input size 416 $\times$ 416), 
 on each of the two datasets.} We trained each network for the number of iterations necessary to its convergence. We notice, for example, in Table~\ref{tab:my-table-experiments} that YOLOv3 necessitated a higher number of iterations when using the largest input size (608 $\times$ 608) on the Stanford dataset, while it reached convergence after much fewer iterations when using the medium input size (416 $\times$ 416) on the same dataset. {Meanwhile, YOLOv4 converges much faster in all configurations due to the use of the cosine annealing scheduler described in Section~\ref{section-yolo-v4}}.  Nevertheless, the number of steps needed to reach convergence is non-deterministic and depends on the initialization of the weights.


\begin{specialtable}[H]
\centering
\caption{Details of the main experiments. The default configuration of Faster R-CNN allows for a variable input size that conserves the aspect ration of the image. In this case, the input size shown is an average.} 
\label{tab:my-table-experiments}
\setlength{\cellWidtha}{\columnwidth/6-2\tabcolsep-0.5in}
\setlength{\cellWidthb}{\columnwidth/6-2\tabcolsep+0.15in}
\setlength{\cellWidthc}{\columnwidth/6-2\tabcolsep+0.15in}
\setlength{\cellWidthd}{\columnwidth/6-2\tabcolsep+0.0in}
\setlength{\cellWidthe}{\columnwidth/6-2\tabcolsep+0.2in}
\setlength{\cellWidthf}{\columnwidth/6-2\tabcolsep+0.0in}
\scalebox{1}[1]{\begin{tabularx}{\columnwidth}{>{\PreserveBackslash\centering}m{\cellWidtha}>{\PreserveBackslash\raggedright\arraybackslash}m{\cellWidthb}>{\PreserveBackslash\raggedright\arraybackslash}m{\cellWidthc}>{\PreserveBackslash\raggedright\arraybackslash}m{\cellWidthd}>{\PreserveBackslash\raggedright\arraybackslash}m{\cellWidthe}>{\PreserveBackslash\raggedright\arraybackslash}m{\cellWidthf}}
\toprule
\multicolumn{1}{c}{\begin{tabular}[c]{@{}c@{}}\textbf{\#}\end{tabular}} & \multicolumn{1}{c}{\textbf{Algorithm}}    & \multicolumn{1}{c}{\begin{tabular}[c]{@{}c@{}}\textbf{Feature}\\  \textbf{Extractor}\end{tabular}} & \multicolumn{1}{c}{\textbf{Dataset}}  & \begin{tabular}[c]{@{}l@{}}\textbf{Average}\\ \textbf{Input Size}\end{tabular} & \begin{tabular}[c]{@{}l@{}}\textbf{Number of}\\ {\textbf{Iterations}}\end{tabular} \\ \midrule
1                                                                              & {Faster R-CNN} & {Inception v2}                                                 & {Stanford} & 816 $\times$ 600 (variable)                                              & 600,000                                                    \\ \midrule
2                                                                              & {Faster R-CNN} & {Inception v2}                                                 & PSU                           & 992 $\times$ 550 (variable)                                              & 600,000                                                    \\ \midrule
3                                                                              & {Faster R-CNN} & Resnet50                                                                          & {Stanford} & 816 $\times$ 600 (variable)                                              & 600,000                                                    \\ \midrule
4                                                                              & Faster R-CNN                      & Resnet50                                                                          & PSU                           & 992 $\times$ 550 (variable)                                              & 600,000                                                    \\ \midrule
5                                                                              & {Faster R-CNN} & {Inception v2}                                                 & {Stanford} & 608 $\times$ 608 (fixed)                                              & 600,000                                                    \\ \midrule
6                                                                              & {Faster R-CNN} & {Inception v2}                                                 & PSU                           & 608 $\times$ 608 (fixed)                                                & 600,000                                                    \\ \midrule
7                                                                              & {Faster R-CNN} & Resnet50                                                                          & {Stanford} & 608 $\times$ 608 (fixed)                                               & 600,000                                                    \\ \midrule
8                                                                              & Faster R-CNN                      & Resnet50                                                                          & PSU                           & 608 $\times$ 608 (fixed)                                                & 600,000                                                    \\ \midrule

9                                                                                 & YOLO v3                           & Darknet-53                                                                        & Stanford                      & 320 $\times$ 320 (fixed)                                              & {896,000}                                                    \\ \midrule                                           10                               & YOLO v3                           & Darknet-53                                                                        & Stanford                      & 416 $\times$ 416 (fixed)                                              & {320,000}                                                    \\ \midrule
11                                                                              & YOLO v3                           & Darknet-53                                                                        & Stanford                      & 608 $\times$ 608 (fixed)                                              & {1,088,000}                                                    \\ \midrule

12                                                                             & YOLO v3                           & Darknet-53                                                                        & PSU                           & 320 $\times$ 320 (fixed)                                              & {640,000}                                                     \\ \midrule                                               13                               & YOLO v3                           & Darknet-53                                                                        & PSU                           & 416 $\times$ 416 (fixed)                                              & {640,000}                                                     \\ \midrule
14                                                                              & YOLO v3                           & Darknet-53                                                                        & PSU                           & 608 $\times$ 608  (fixed)                                             & {640,000}                                                     \\ \midrule
{15}                                                                                 & {YOLO v4}                           & {CSPDarknet-53}                                                                        & {Stanford}                      & {320 $\times$ 320 (fixed)}                                              & {192,000}                                                    \\ \midrule                                           {16}                               & {YOLO v4}                           & {CSPDarknet-53}                                                                        & {Stanford}                      & {416 $\times$ 416 (fixed)}                                              & {192,000}                                                    \\ \midrule
{17}                                                                              & {YOLO v4}                           & {CSPDarknet-53}                                                                        & {Stanford}                      & {608 $\times$ 608 (fixed)}                                              & {192,000}                                                    \\ \midrule

{18}                                                                             & {YOLO v4}                           & {CSPDarknet-53}                                                                        & {PSU}                           & {320 $\times$ 320 (fixed)}                                              & {192,000}                                                     \\ \midrule                                               {19}                               & {YOLO v4}                           & {CSPDarknet-53}                                                                        & {PSU}                           & {416 $\times$ 416 (fixed)}                                              & {192,000}                                                     \\ \midrule
{20}                                                                              & {YOLO v4}                           & {CSPDarknet-53}                                                                        & {PSU}                           & {608 $\times$ 608  (fixed})                                             & {192,000}                                                     \\ \bottomrule

\end{tabularx}}
\end{specialtable}

\subsection{Results and Discussion}\label{sec:results}
For the experimental setup, we used a workstation powered by an Intel core i7-8700K (3.7 GHz) processor, with 32 GB RAM, and an NVIDIA GeForce 1080 (8 GB) GPU, running on Linux. {We will first explain the metrics used for the evaluation, then discuss the results of each metric for each algorithm on each testing dataset described above. We also tested different learning rates and anchor scales in order to assess the algorithms' sensitivity to these hyperparameters. A total of 52 trainings have been conducted (20 experiments with default hyperparameters, 28 experiments with different learning rates, and 4 experiments with different anchor scales).}

{\subsubsection{{Metrics}}}

The following metrics have been used to assess the results:

\begin{itemize}
    \item IoU: Intersection over Union measuring the overlap between the predicted and the ground-truth bounding boxes.
    \item mAP: mean average precision, or simply AP, since we are dealing with only one class. It corresponds to the area under the precision vs. recall curve. AP was measured for different values of IoU (0.5, 0.6, 0.7, 0.8, and 0.9).
    \item FPS: number of frames per second, measuring the inference processing speed.
    \item Inference time (in millisecond per image): also measuring the processing speed. \[{Inference\ time\ (ms) = \frac{1000}{FPS}}\]
    \item AR\textsuperscript{max=1}, AR\textsuperscript{max=10}, and AR\textsuperscript{max=100}: average recall, when considering a maximum number of detections per image, averaged over all values of IoU specified above. {We allow only the 1, 10, or 100 top-scoring detections for each image. This metric penalizes missing detections (false negatives) and duplicates (several bounding boxes for a single object).}
\end{itemize}


\subsubsection{{Average Precision}}
When analyzing the results, it appears that {all three tested algorithms} gave a much better AP on the PSU dataset than on the Stanford dataset (Figure~\ref{fig:Average_AP}). This is mainly due to the fact that, contrary to the PSU dataset, the characteristics of the Stanford dataset differ largely between the training and testing images, as detailed in IV.A
. This is the well known problem of domain adaptation in machine learning~\cite{Benjdira2019Segmentation}. The Stanford dataset contains 20~times more car instances than the PSU dataset (Tables \ref{tab:Stanford_and_PSU_Datasets}), whereas the performance of Faster R-CNN, YOLOv3{, and YOLOv4 algorithms} was respectively four, seven, and five times better on the PSU dataset, in terms of AP. This highlights the fact that the clarity of the features, the quality of annotation{, and the representativity of the learning dataset are more important than the actual size of the dataset.  }

\begin{figure}[H]
\includegraphics[width=10cm]{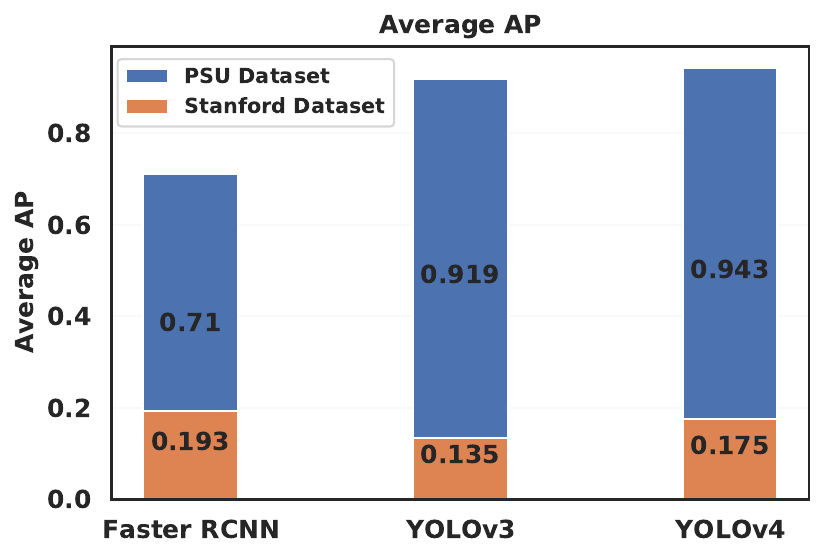}
\caption{Comparison of the AP (Average Precision) 
 between YOLOv3{, YOLOv4,} and Faster R-CNN.
\label{fig:Average_AP}}  
\end{figure}

However, Figure~\ref{fig:FP_TP_FN} shows that the number of false negatives (non-detected cars) is much higher than the number of false positives on the Stanford dataset (3 times higher for Faster R-CNN,    73 times higher for YOLOv3{, and 66 times higher for YOLOv4}), and   much higher than the number of true positives, which indicates that most cars go undetected in the Stanford dataset, {most likely due to the different size and aspect ratio of  the cars in the testing images, compared to the training images.} {This is also visible on \mbox{Figure~\ref{Precision_vs_Recall_Stanford}}, which illustrates the trade-off between precision and recall for different score thresholds. While the precision is close to 1 for YOLOv3 and YOLOv4, but significantly lower for Faster R-CNN, all the algorithms have a recall inferior to 0.25 on the Stanford dataset. On the contrary, Figure~\ref{Precision_vs_Recall_PSU} shows high values of recall for YOLOv3 and YOLOv4, and a slightly lower precision compared to Faster R-CNN, on the PSU dataset.}
Even though {all three} algorithms performed poorly on the Stanford dataset as compared to  the PSU dataset, with less than 20\% of AP, there is still a statistically significant difference between Faster R-CNN and YOLOv3 on this dataset. In fact, a T-test between the two sets of AP values of the two algorithms (for different IoU and score thresholds) yielded a p-value of 0.0020, which means that the null hypothesis (equality of the means of the two sets of AP values) can be rejected with a confidence of 99.8\%. {Meanwhile, the p-value between YOLOv3 and YOLOv4 AP values is 0.72, which means that the difference in performance between these two algorithms is not statistically significant, as opposed to the large improvement that Bochkovskiy et al.~\cite{yolov4} obtained on the  COCO dataset. This result may indicate that YOLOv4 has been specifically tuned for the COCO dataset and does not perform as well on other datasets in terms of AP.}

\end{paracol}
\nointerlineskip
\begin{figure}[H]
\widefigure
\includegraphics[width=17cm]{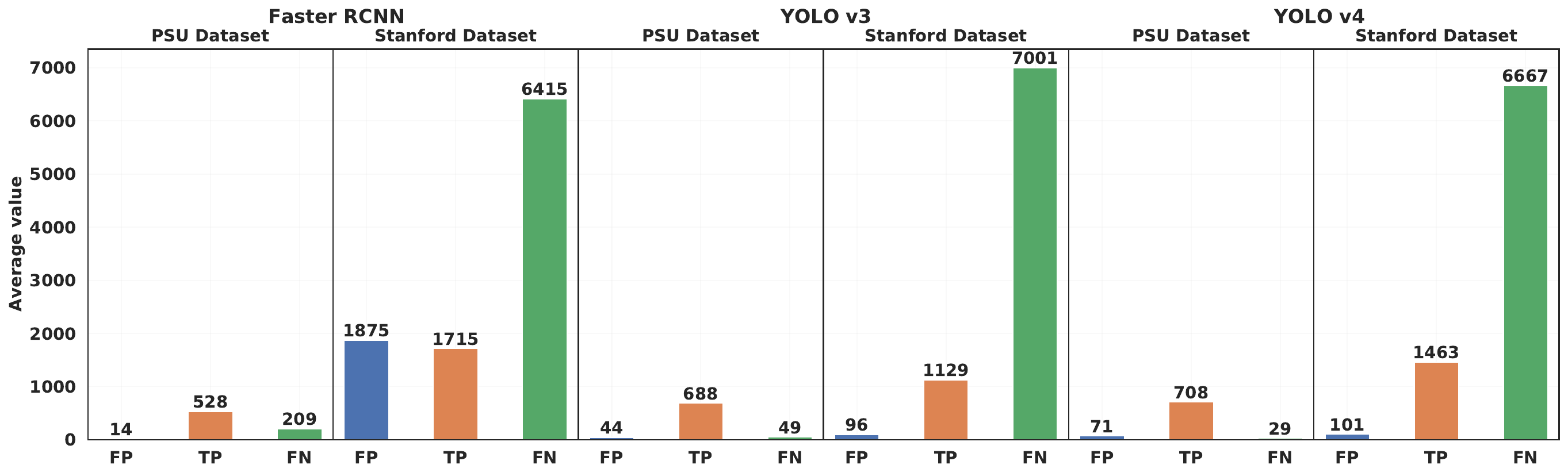}
\caption{Average number of false positives (FP), false negatives (FN), and true positives (TP) for YOLOv3{, YOLOv4,} and Faster R-CNN on the two datasets.
\label{fig:FP_TP_FN}}  
\end{figure}
\begin{paracol}{2}
\switchcolumn
\vspace{-6pt}

\begin{figure}[H]
\includegraphics[width=12cm]{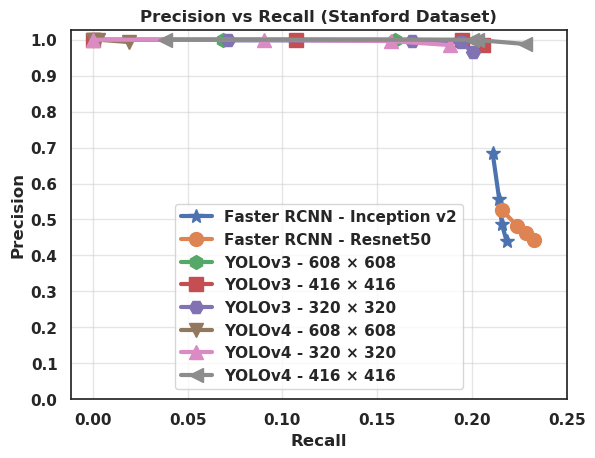}
\caption{{Precision vs. Recall for different values of score threshold (0.3, 0.5, 0.7, and 0.9), and IoU = 0.6 (Intersection over Union), on the Stanford dataset.} 
\label{Precision_vs_Recall_Stanford}}  
\end{figure}

\begin{figure}[H]
\includegraphics[width=12cm]{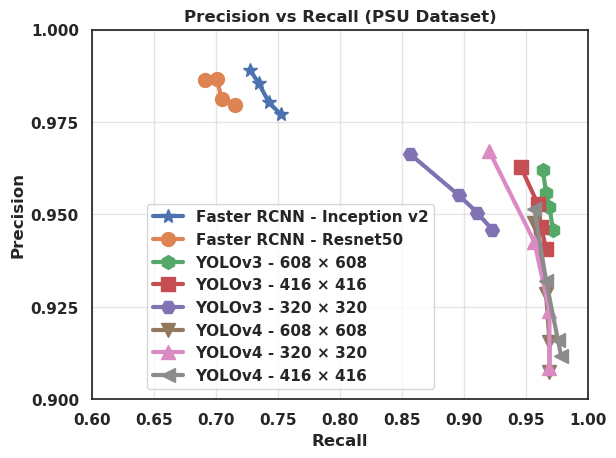}
\caption{{Precision vs. Recall for different values of score threshold (0.3, 0.5, 0.7, and 0.9), and IoU = 0.6 (Intersection over Union), on the PSU dataset.}
\label{Precision_vs_Recall_PSU}}  
\end{figure}

Figure~\ref{fig:Yolo_Stanford_misclassif} shows examples of YOLOv3 and Faster R-CNN misclassifications on a sample image of the Stanford dataset. The false positives shown may be explained by the presence of errors of annotations in the learning dataset, as mentioned in Section~\ref{section-datasets}. Figures~\ref{fig:Yolo_PSU_Misclassif} and~\ref{fig:FRCNN_PSU_Misclassif} show examples of YOLOv3 and Faster R-CNN misclassifications (all of them false negatives) on a sample image of the PSU dataset, respectively. {YOLOv4 yields almost equivalent misclassifications, compared to YOLOv3.}

\begin{figure}[H] 
\includegraphics[width=13.5cm]{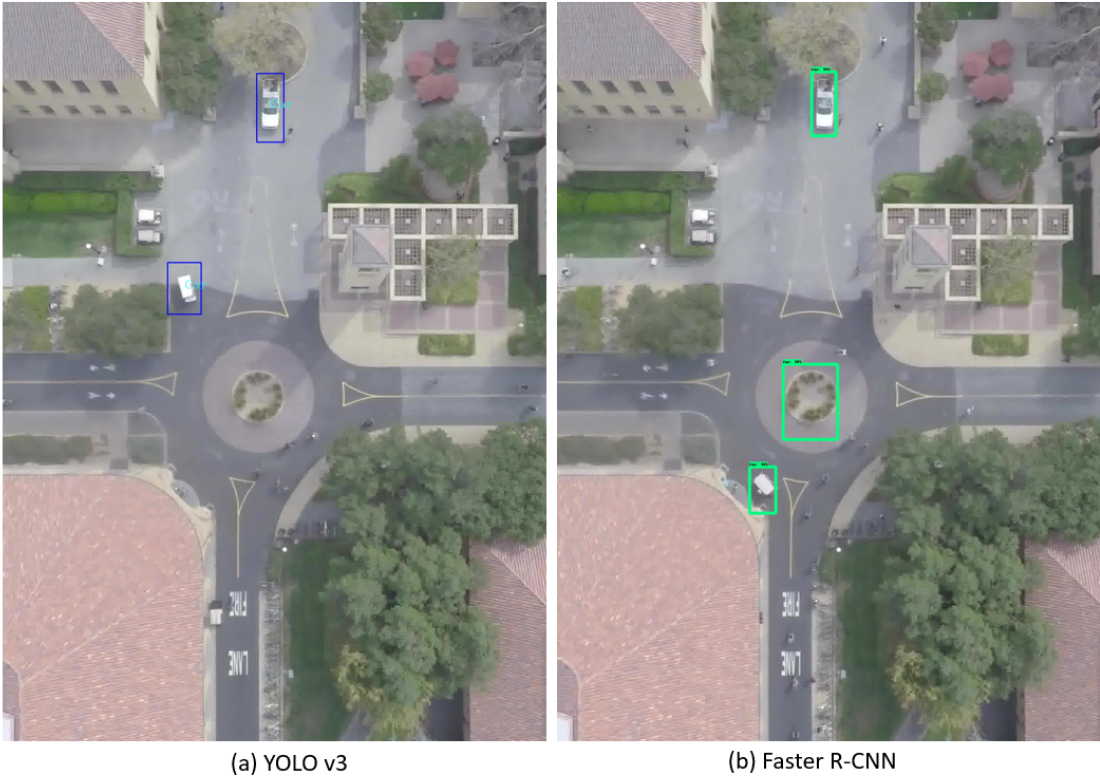}
\caption{\textls[-45]{Example of (\textbf{a}) YOLOv3 and (\textbf{b}) Faster R-CNN's 
 output on a sample image of the Stanford dataset.}}  \label{fig:Yolo_Stanford_misclassif}
\end{figure}

\begin{figure}[H]
\includegraphics[width=13cm]{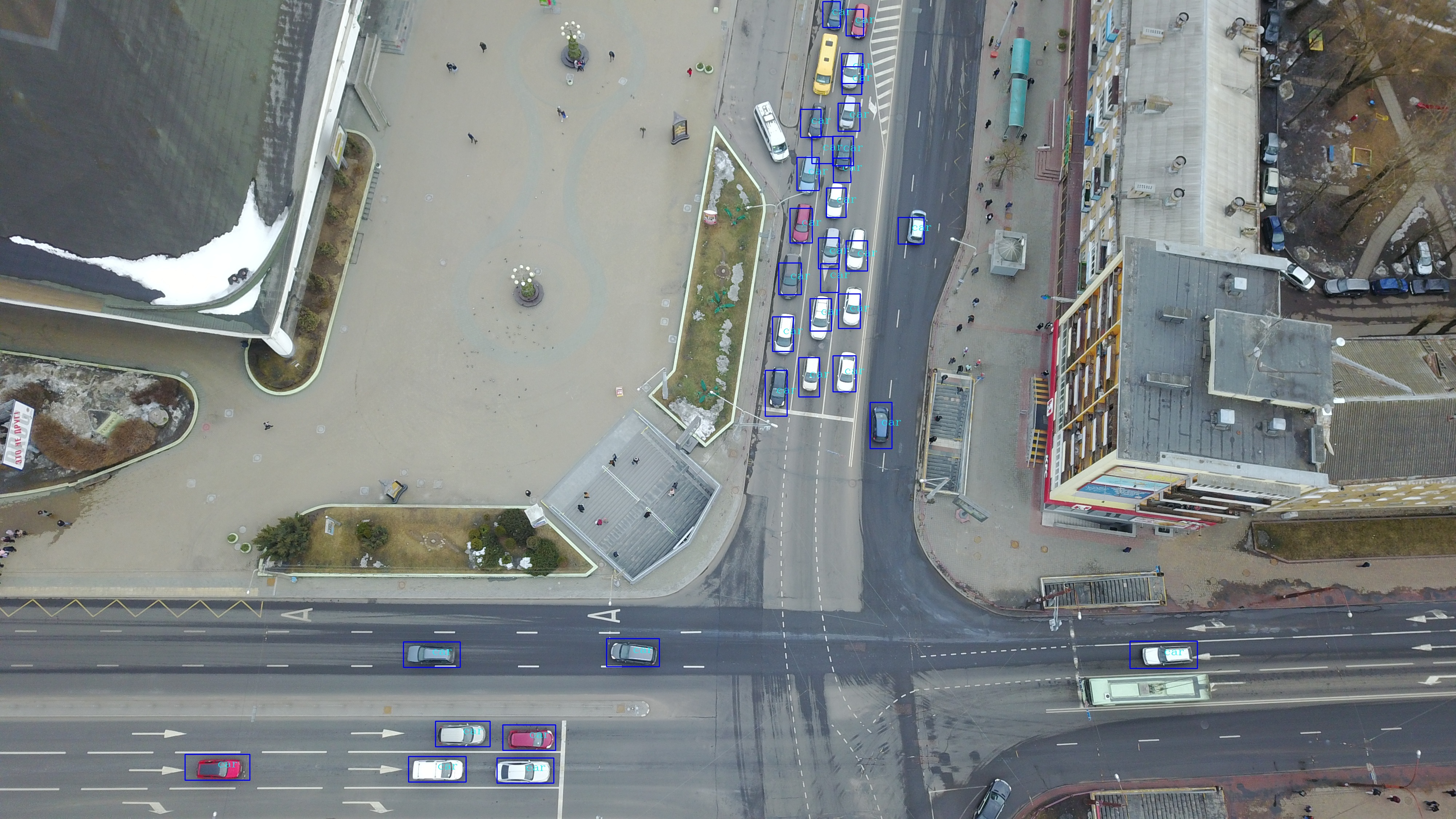}
\caption{Example of YOLOv3's output on an image of the PSU 
 dataset, showing a few false negatives (non-detected cars).
\label{fig:Yolo_PSU_Misclassif}}  
\end{figure}
\vspace{-6pt}

\begin{figure}[H]
\includegraphics[width=13cm]{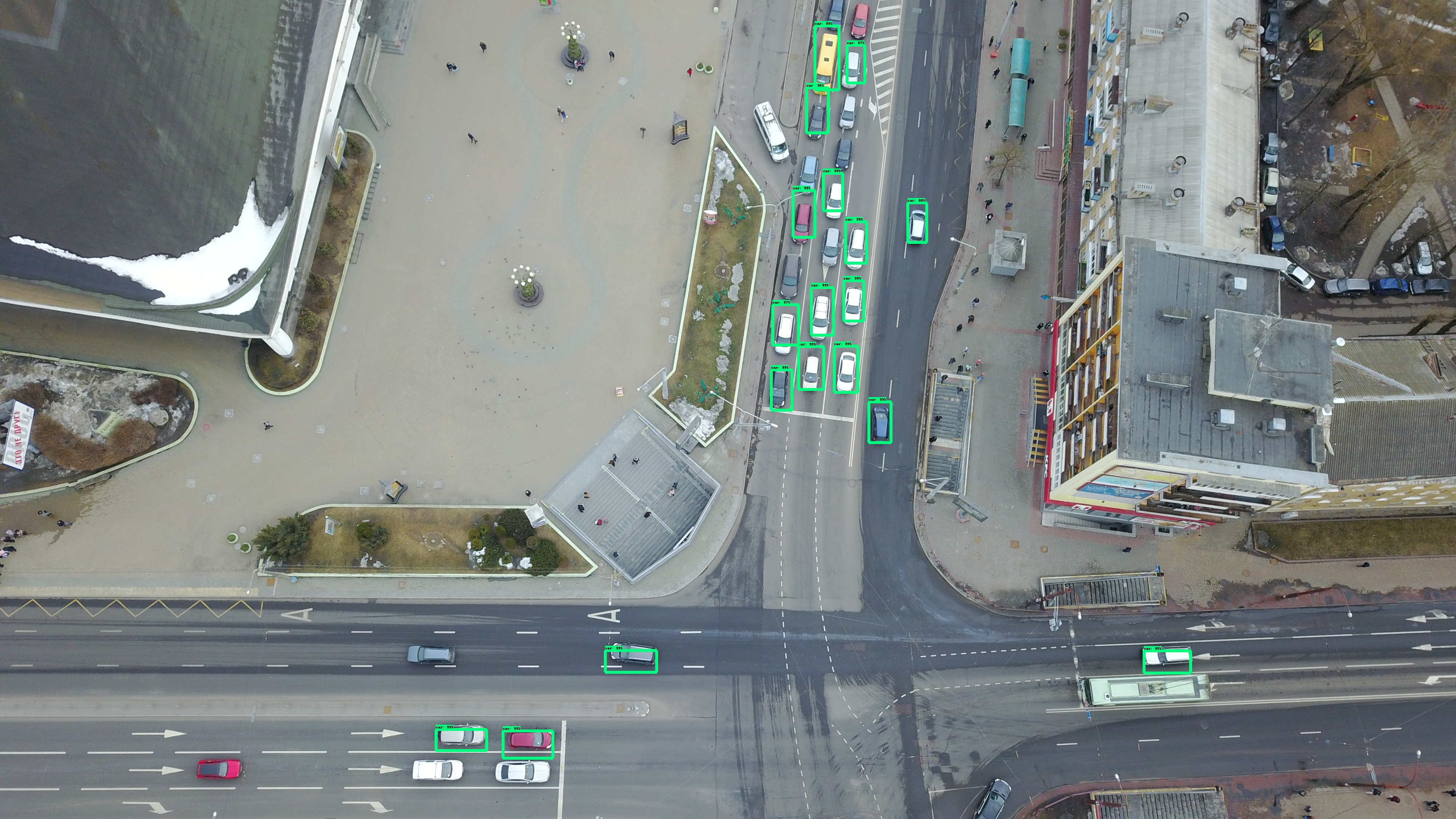}
\caption{Example of Faster R-CNN misclassifications on an image of the PSU 
 dataset, showing several false negatives (non-detected cars).
\label{fig:FRCNN_PSU_Misclassif}}  
\end{figure}
\vspace{-6pt}


\subsubsection{{Average Recall}}
Table~\ref{tab:AR_Stanford} shows the average  recall  for  a  given  maximum  number  of detections {(as described in the introduction of Section~\ref{sec:results})}, on the Stanford dataset. {YOLOv4 (with medium and high input size) shows the best results in this metric, while the small input size (320 $\times$ 320) shows a marked inferior performance for both YOLOv3 and YOLOv4.} The fact that the columns AR\textsuperscript{max=10} and  AR\textsuperscript{max=100} in this table are identical can be explained by the fact that very few images in the Stanford testing dataset contain more than 10 car instances. {Nevertheless, we have kept this duplicated column to compare it to} Table~\ref{tab:AR_PSU}, which shows the same metrics on the PSU dataset. {YOLOv4 (with any input size) is significantly better in terms of the three metrics on this dataset, which indicates that YOLOv4} is better at detecting a high number of objects in a single image.

\begin{specialtable}[H]
\caption{Average 
recall for a given maximum number of detections, averaged over all values of IoU (0.5, 0.6, 0.7, 0.8, and 0.9) (Intersection over Union), on the Stanford dataset. The best results are marked in bold.} 
\label{tab:AR_Stanford}
\begin{tabular*}{\hsize}{@{}@{\extracolsep{\fill}}llll@{}}
\toprule
\textbf{Network}                     & \textbf{AR\textsuperscript{max=1}}             & \textbf{AR\textsuperscript{max=10}}            & \textbf{AR\textsuperscript{max=100}}           \\ \midrule
Faster R-CNN (Inception-v2) & 15.1\%          & 17.1\%          & 17.1\%          \\ 
Faster R-CNN (Resnet50)     & 16.4\% & 18.6\% & 18.6\% \\ 
YOLOv3 (320 $\times$ 320)            & 9.0\%         & 9.1\%         & 9.1\%         \\ 
YOLOv3 (416 $\times$ 416)            & 17.1\%          & 17.3\%          & 17.3\%          \\ 
YOLOv3 (608 $\times$ 608)            & 17.2\%           & 17.3\%           & 17.3\%            \\ 
{YOLOv4 (320 $\times$ 320)}            & {14.7\%}         & {14.7\%}         & {14.7\%}         \\ 
{YOLOv4 (416 $\times$ 416)}            & {\textbf{19.3\%}}         & {19.4\%}          & {19.4\%}          \\ 
{YOLOv4 (608 $\times$ 608)}            & {19.1\%}           & {\textbf{24.0\%}}           & {\textbf{24.0\%}}            \\ \bottomrule      
\end{tabular*}  
\end{specialtable}
\vspace{-6pt}

\begin{specialtable}[H]
\centering
\caption{Average 
recall for a given maximum number of detections, averaged over all values of IoU (0.5, 0.6, 0.7, 0.8, and 0.9), on the PSU dataset. The best results are marked in bold.} 
\label{tab:AR_PSU}
\begin{tabular*}{\hsize}{@{}@{\extracolsep{\fill}}llll@{}}
\toprule
\textbf{Network}                     & \textbf{AR\textsuperscript{max=1}}             & \textbf{AR\textsuperscript{max=10}}            & \textbf{AR\textsuperscript{max=100}}           \\ \midrule
Faster R-CNN (Inception-v2) & 6.2\%          & 41.5\%          & 70.8\%          \\ 
Faster R-CNN (Resnet50)     & 6.4\% & 41.5\% & 67.2\% \\ 
YOLOv3 (320 $\times$ 320)            & 6.0\%         & 42.2\%         & 81.0\%         \\ 
YOLOv3 (416 $\times$ 416)            & 6.4\%          & 44.1\%          & 90.4\%          \\ 
YOLOv3 (608 $\times$ 608)            & 6.4\%           & 44.5\%           & 91.9\%            \\ 
{YOLOv4 (320 $\times$ 320)}            & {\textbf{6.8\%}}         & {\textbf{47.1\%}}         & {95.5\%}         \\ 
{YOLOv4 (416 $\times$ 416)}            & {\textbf{6.8\%}}          & {46.8\%}          & {\textbf{96.6\%}}          \\ 
{YOLOv4 (608 $\times$ 608)}            & {6.7\%}           & {46.5\%}           & {95.6\%}            \\ \bottomrule     
\end{tabular*}  
\end{specialtable}

{\subsubsection{{Inference Speed}}\label{sec:inference_speed}}

{Figure~\ref{fig:FPS_by_InputSize} depicts the inference speed measured in frames per second (FPS), for each of the tested algorithms on both datasets. It shows that all configurations of YOLOv3 and YOLOv4 are significantly faster than Faster R-CNN. Moreover, the input size has a direct impact on the inference time, as expected, since a larger input size generates a greater number of network parameters, and hence a larger number of operations. In fact, the inference processing speed of both YOLOv3 and YOLOv4 largely depends on the input size (from 12 FPS for 608 $\times$ 608 up to 23 FPS for 320 $\times$ 320),
 with little variation between the two datasets. As for Faster R-CNN, the Inception v2 feature extractor is 2.3 and 1.5 times faster on the Stanford and PSU datasets, respectively. The difference in speed when applying these algorithms 
 on the two datasets is explained by the difference of image input size. In fact, we calculated that the average number of pixels in  the input test images (after resizing) is 544,000 for  the PSU dataset, and 265,000 for  the Stanford dataset, whereas YOLOv3 and YOLOv4 are not affected by this difference because they
 resize the images to a fixed input size. }

{The inference speed of YOLOv3 and YOLOv4 is nearly real-time. Nevertheless, if we want to run these object detectors on embedded edge devices on UAVs, which have reduced capabilities compared to the GPU workstation used here, we should apply model   optimizations after training, as explained in~\cite{koubaa2021cloud}. 
}
\clearpage
\end{paracol}
\nointerlineskip
\begin{figure}[H]
\widefigure
\includegraphics[width=16cm,height=7cm]{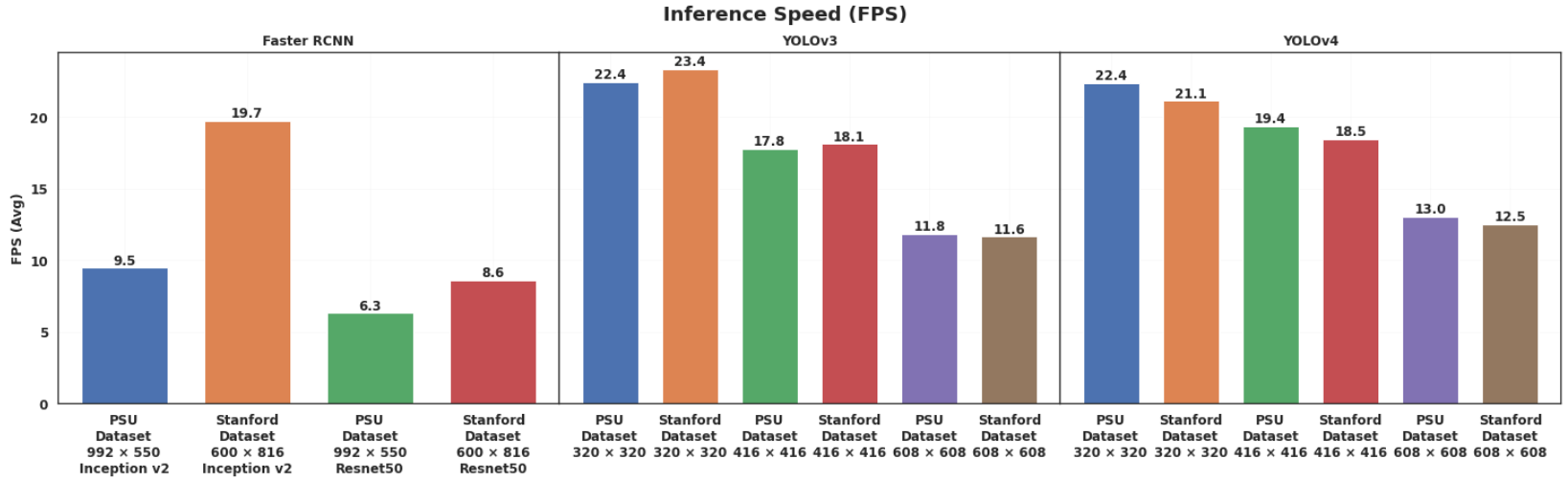}
\caption{Inference speed measured in frames per second (FPS), for each of the tested algorithms. {The input size for YOLOv3 and YOLOv4 is fixed, whereas the value shown for Faster R-CNN is an average of the variable input sizes.} 
\label{fig:FPS_by_InputSize}}  
\end{figure}
\begin{paracol}{2}
\switchcolumn
\vspace{-6pt}

\subsubsection{{Effect of the Dataset Characteristics}}
YOLOv3 {(and to a slightly lesser extent YOLOv4)} show the largest performance discrepancy between the two datasets. While they provide a very high recognition on the PSU dataset (up to 0.965 of AP), {their} performance {markedly} decreases on the Stanford dataset (Figure~\ref{fig:Average_AP}). 
This is mainly due to the spatial constraints imposed by {the YOLO family of algorithms}. On the other hand, Faster R-CNN was designed to better deal with objects of various scales and aspect ratios~\cite{Faster_R-CNN_journal}. 

Nevertheless, the contrary can be observed in terms of IoU (Figure~\ref{fig:IoU}). While the average IoU of Faster R-CNN decreases by half between  the PSU dataset and the  Stanford dataset, it decreases only by {9\% for YOLOv4 and 11\% for YOLOv3}. The imprecision of the ground-truth bounding boxes in the Stanford dataset {and the discrepancy between training and testing features could explain the difference between the two datasets in terms of IoU.} {YOLOv4 and }YOLOv3, however, manage to keep relatively precise predicted bounding boxes on both datasets. {YOLOv4 shows the best average IoU on the Stanford dataset, due to its use of the CIoU loss function, as explained in Section~\ref{section-yolo-v4}.}

\begin{figure}[H]
\includegraphics[width=10cm]{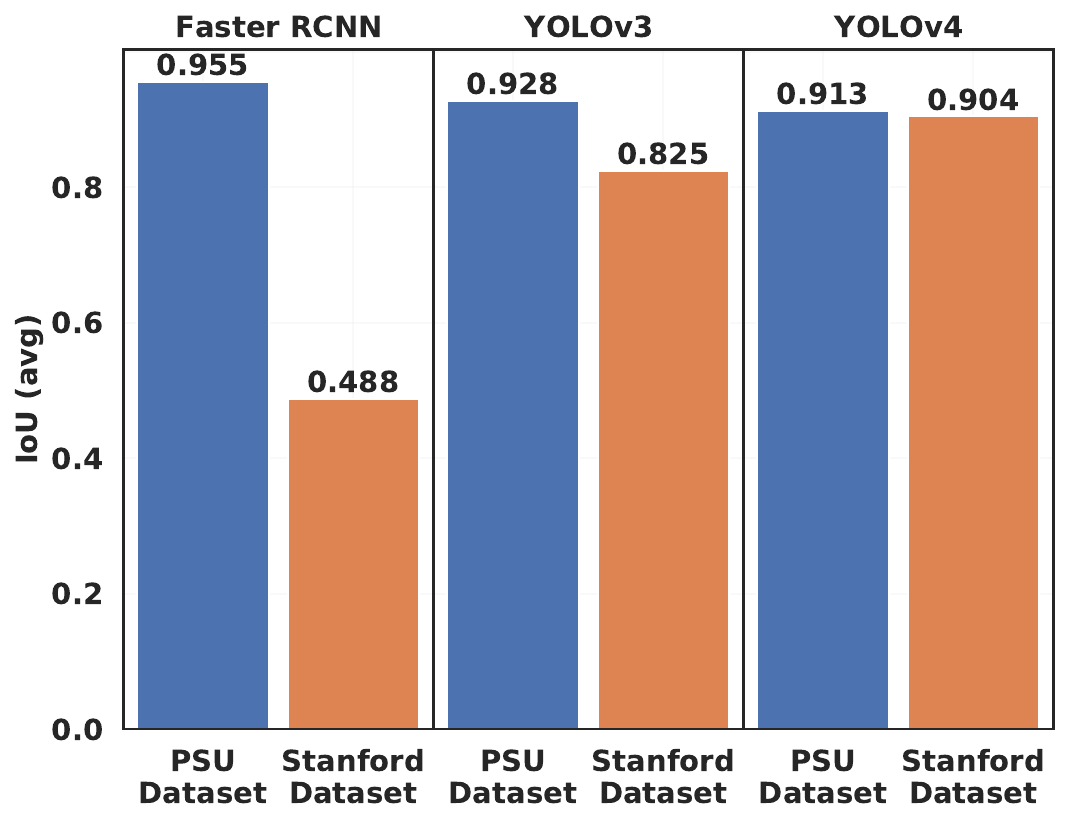}
\caption{Average IoU (Intersection over Union) 
 value for YOLOv3{, YOLOv4,} and Faster R-CNN 
  on the two datasets.
\label{fig:IoU}}  
\end{figure}

In addition, Faster R-CNN shows a high disparity between the two datasets in terms of processing speed (2.7 times faster on the Stanford dataset), mainly due to the difference in image input size{, as mentioned in Section~\ref{sec:inference_speed}}. 

\subsubsection{{{Effect of Object Size}}}
{Figures \ref{fig:AP_by_size_PSU} and \ref{fig:AP_by_size_Stanford} show the Average Precision (AP) for each category of object size on the PSU and Stanford datasets respectively. We define small objects as objects having a surface less than 5000 pixel\textsuperscript{2}, medium objects as having a surface between 5000 and 10,000 pixel\textsuperscript{2}, and large objects as having a surface greater than 10,000 pixel\textsuperscript{2}. 
We notice that the pattern is the same for all the tested networks. On the PSU dataset, the best performance is always obtained on small objects, whereas the lowest performance is obtained for medium-size objects (with the exception of Faster R-CNN/Resnet50 that exhibits a slightly lower AP for large objects). By contrast, on the Stanford dataset, all the algorithms completely fail to detect small and medium-size cars, while showing a much better performance on large objects.
In both cases, this can be explained by the distribution of car sizes in the training dataset (Figure~\ref{fig:hist_car_sizes}). In fact, in the PSU training dataset, the category of small cars is the most well represented (87\% of all objects), while the category of medium-size and large cars are much less represented (8\% and 4\%, respectively). On the other hand, in the Stanford training dataset, the most represented category is large cars (58\%), while small and medium-size cars are less represented (5\% and 38\%, respectively). In addition, large objects still have the additional advantage of possessing more discernible features,  hence being easier to~detect.
}
\begin{figure}[H]
\includegraphics[width=11cm]{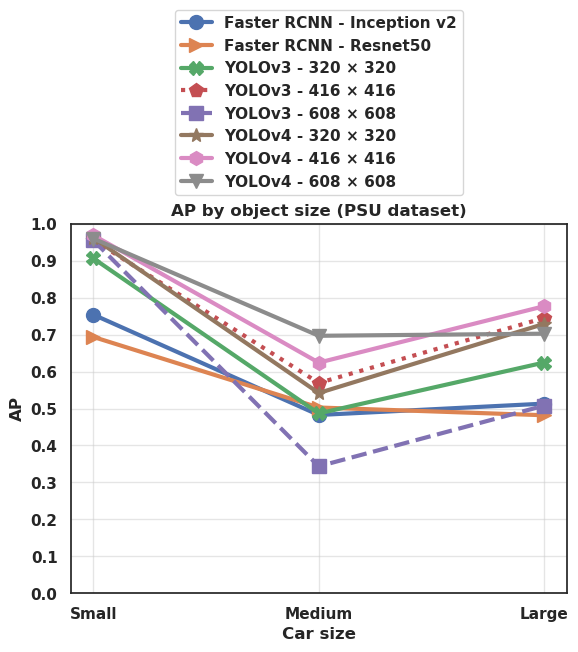}
\caption{{Average Precision (AP) for each 
 category of object size: small (object surface $<$ 5000 pixel\textsuperscript{2}), medium-size (5000 pixel\textsuperscript{2} $\leq$ object surface $\leq$ 10,000 pixel\textsuperscript{2}), and large (object surface $>$ 10,000 pixel\textsuperscript{2}), on the PSU dataset.} 
\label{fig:AP_by_size_PSU}}  
\end{figure}

\begin{figure}[H]
\includegraphics[width=12cm]{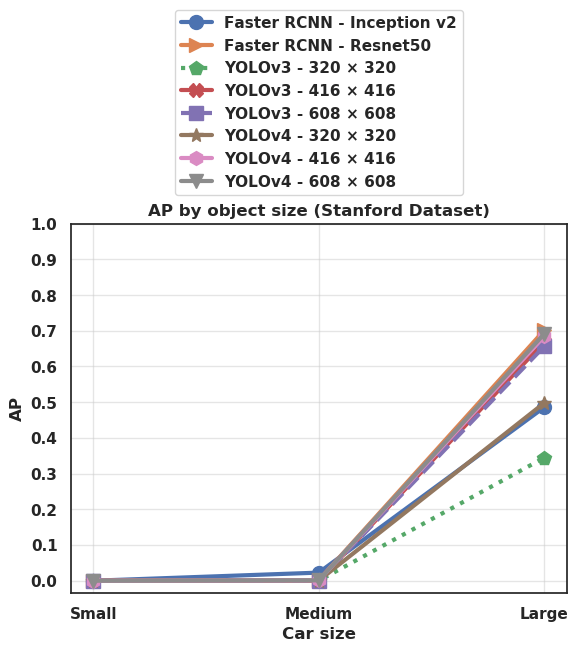}
\caption{{AP for each category of object 
 size: small (object surface $<$ 5000 pixel\textsuperscript{2}), medium-size (5000   pixel\textsuperscript{2} $\leq$ object surface $\leq$ 10,000 pixel\textsuperscript{2}), and large (object surface $>$ 10,000 pixel\textsuperscript{2}), on the Stanford dataset.} 
\label{fig:AP_by_size_Stanford}}   
\end{figure}

\subsubsection{{Effect of the Feature Extractor}}
The effect of the feature extractor for Faster R-CNN is very limited on the AP, except for a high value of IoU threshold (0.9) on the Stanford dataset, as can be seen in \mbox{Figure~\ref{fig:ap_iou_psu} and~\ref{fig:ap_iou_stanford}}. Nevertheless, in terms of inference speed, the Inception-v2 feature extractor is significantly faster than Resnet50 (Figures \ref{fig:AP_time_PSU} and \ref{fig:AP_time_Stanford}), which is consistent with the findings of Bianco et al.~\cite{bianco2018benchmark} who also showed that Inception-v2 (also known as BN-inception) is less computationally complex.

\subsubsection{{Effect of the Input Size}}
Figures \ref{fig:AP_time_PSU} and \ref{fig:AP_time_Stanford} show a significant gain in YOLOv3's AP when moving from a \mbox{320 $\times$ 320} input size to 416 $\times$ 416, but the performance stagnates when we move further to 608 $\times$ 608, which means that the 416 $\times$ 416 resolution is sufficient to detect the objects of the two datasets, and a higher input size may lead to overfitting. {A similar behavior can be observed for YOLOv4, except that the improvement between 320 $\times$ 320 and 416 $\times$ 416 sizes is much lower on the PSU dataset, since the first input size already provides   an excellent AP. Moreover, we observe a decrease in AP, when we move to 608 $\times$ 608 on the PSU dataset. This reveals an over-fitting on this smaller dataset, when using more complex networks.}  Concerning Faster R-CNN, Tables~\ref{tab:Detailed_results_PSU} {and \ref{tab:Detailed_results_Stanford}} show that the default variable input size, which conserves the aspect ratio of the images, provides a better precision and recall than the fixed size configuration, in all cases except with Inception-v2 on the Stanford dataset, which results in significantly fewer false negatives (5215 compared to 6351). This is likely due to an exceptional congruence between the fixed input size and the anchor scales for Inception-v2 on this particular dataset. This configuration also gives a slightly better performance in terms of inference speed (21.1 FPS compared to 19.2 FPS), due to the smaller average input size. {In fact, the image input size has a direct impact on the inference speed, as explained in Section~\ref{sec:inference_speed}.}

\begin{figure}[H]
\includegraphics[width=12cm]{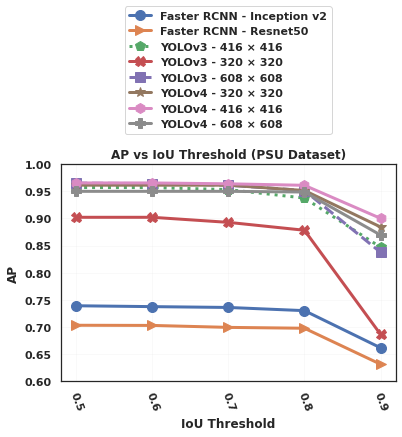}
\caption{\textls[-45]{AP (Average Precision), at different IoU (Intersection over Union) 
 threshold values, of the tested algorithms on the PSU 
 dataset.}
\label{fig:ap_iou_psu}}  
 \end{figure}

 \begin{figure}[H]
\includegraphics[width=12cm]{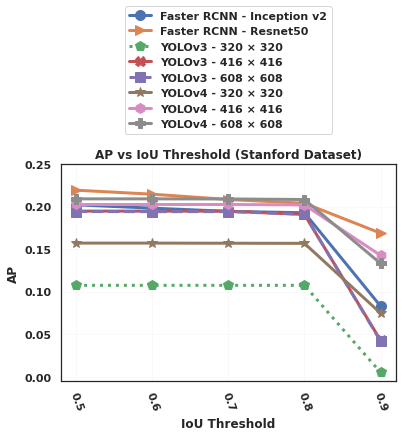}
\caption{AP (Average Precision), at different IoU (Intersection over Union) 
 threshold values, of the tested algorithms on the Stanford~dataset.
\label{fig:ap_iou_stanford}}
\end{figure}

\begin{figure}[H]
\includegraphics[width=13cm]{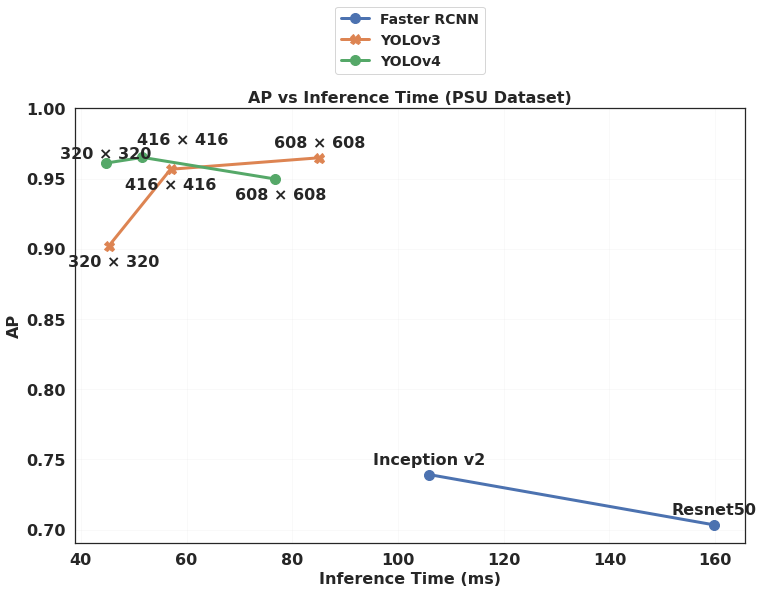}
\caption{Comparison of the trade-off between AP (Average Precision)  
and inference time for {YOLOv4 and} YOLOv3 (with 3 different input sizes each) and for Faster R-CNN 
 (with two different feature extractors), on the PSU 
 dataset.
\label{fig:AP_time_PSU}}  
\end{figure}
\vspace{-6pt}

\begin{figure}[H]
\includegraphics[width=13cm]{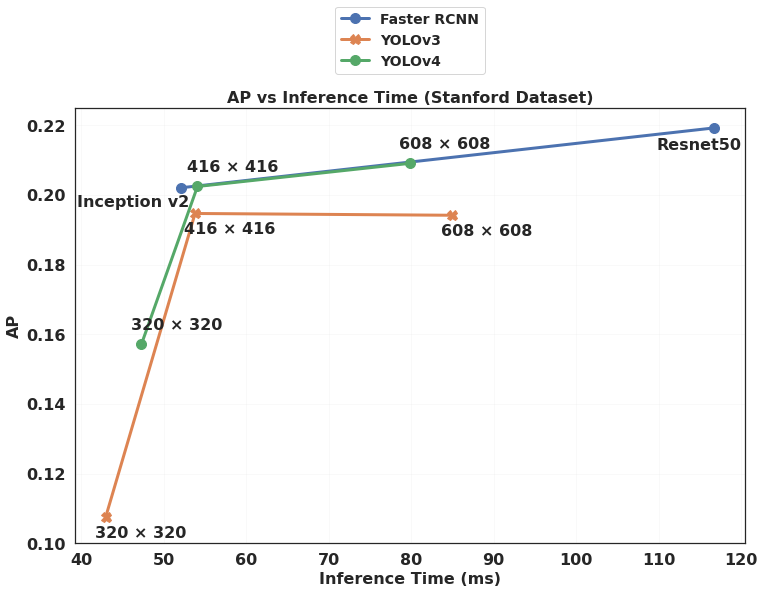}
\caption{Comparison of the trade-off between AP (Average Precision) 
and inference time for {YOLOv4 and} YOLOv3 (with 3 different input sizes each) and for Faster R-CNN 
 (with two different feature extractors), on the Stanford dataset.
\label{fig:AP_time_Stanford}}  
\end{figure}

\clearpage
\end{paracol}
\nointerlineskip
\begin{specialtable}[H]
\widetable
\caption{Detailed 
results of different configurations of YOLOv3, YOLOv4, and Faster R-CNN, on  the PSU 
 dataset. The default configuration of Faster R-CNN allows for a variable input size that conserves the aspect ratio of the image. In this case, the input size shown is an average. The best results are shown in bold.} 
\label{tab:Detailed_results_PSU}
\resizebox{\textwidth}{!}{\begin{tabular}{lllllllllllc}
\toprule
\textbf{Algorithm} & \multicolumn{1}{l}{\textbf{\begin{tabular}[c]{@{}l@{}}Feature \\ Extractor\end{tabular}}} & \textbf{Input Size} & \textbf{{AP}} & \textbf{TP}   & \textbf{FN}   & \textbf{FP} & \textbf{Precision} & \textbf{Recall} & \multicolumn{1}{l}{\textbf{\begin{tabular}[c]{@{}l@{}}F1 \\ Score\end{tabular}}} & \textbf{FPS} & \multicolumn{1}{c}{\textbf{\begin{tabular}[c]{@{}c@{}}{Inference} \\ {Time (ms)}\end{tabular}}}\\ \midrule

Faster R-CNN        & Inception v2               & 992 $\times$ 550 (variable)            & {0.739}              & 548           & 190           & 11          & 0.980              & 0.743           & 0.845            & 9.5 & {105}\\ 
Faster R-CNN        & Inception v2               & 608 $\times$ 608 (fixed)            & {0.731}              & 541           & 197           & 14          & 0.975              & 0.733           & 0.837            & 9.5 & {105}\\ 
Faster R-CNN        & Resnet50                   & 992 $\times$ 550 (variable)             & {0.708}              & 524           & 214           & \textbf{9}  & \textbf{0.983}     & 0.710           & 0.825           & 6.4  & {156}\\ 
Faster R-CNN        & Resnet50                   & 608 $\times$ 608 (fixed)             & {0.623}              & 463           & 275           & 17  & 0.965     & 0.627           & 0.76           & 5.3  & {189}\\
YOLOv3             & Darknet-53                 & 320 $\times$ 320 (fixed)             & {0.902}              & 672           & 66            & 35          & 0.950              & 0.911           & 0.930           & 22.1  & {45}\\ 
YOLOv3             & Darknet-53                 & 416 $\times$ 416 (fixed)             & {0.957}              & 710           & 28            & 40          & 0.947              & 0.962           & 0.954           & 17.5  & {57}\\ 
YOLOv3             & Darknet-53                 & 608 $\times$ 608 (fixed)             & \textbf{{0.965}}              & 715  & \textbf{23} & 36          & 0.952              & 0.969  & \textbf{0.960}    & 11.8 & {84}\\ 
\multicolumn{1}{l}{{YOLOv4}} & \multicolumn{1}{l}{{CSPDarknet-53}} & \multicolumn{1}{l}{{320 $\times$ 320 (fixed)}} & \multicolumn{1}{l}{{0.961}} & \multicolumn{1}{l}{{715}} & \multicolumn{1}{l}{{\textbf{23}}} & \multicolumn{1}{l}{{59}} & \multicolumn{1}{l}{{0.924}} & \multicolumn{1}{l}{{0.969}} & \multicolumn{1}{l}{{0.946}} & \multicolumn{1}{l}{{\textbf{22.4}}} & {\textbf{45}}\\
\multicolumn{1}{l}{{YOLOv4}} & \multicolumn{1}{l}{{CSPDarknet-53}} & \multicolumn{1}{l}{{416 $\times$ 416 (fixed)}} & \multicolumn{1}{l}{{\textbf{0.965}}} & \multicolumn{1}{l}{{\textbf{720}}} & \multicolumn{1}{l}{{18}} & \multicolumn{1}{l}{{66}} & \multicolumn{1}{l}{{0.916}} & \multicolumn{1}{l}{{\textbf{0.976}}} & \multicolumn{1}{l}{{0.945}} & \multicolumn{1}{l}{{19.4}} & {52}\\ 
\multicolumn{1}{l}{{YOLOv4}} & \multicolumn{1}{l}{{CSPDarknet-53}} & \multicolumn{1}{l}{{608 $\times$ 608 (fixed)}} & \multicolumn{1}{l}{{0.950}} & \multicolumn{1}{l}{{715}} & \multicolumn{1}{l}{{\textbf{23}}} & \multicolumn{1}{l}{{66}} & \multicolumn{1}{l}{{0.915}} & \multicolumn{1}{l}{{0.969}} & \multicolumn{1}{l}{{0.941}} & \multicolumn{1}{l}{{13}}  & {77} \\ \bottomrule
\end{tabular}}
\end{specialtable}
\vspace{-6pt}

\begin{specialtable}[H]
\widetable
\caption{Detailed 
results of different configurations of YOLOv3, YOLOv4, and Faster R-CNN, on Stanford dataset. The default configuration of Faster R-CNN
 allows a variable input size that conserves the aspect ration of the image. In this case, the input size shown is an average. The best results are shown in bold.} 
\label{tab:Detailed_results_Stanford}
\resizebox{\textwidth}{!}{\begin{tabular}{lllllllllllc}
\toprule
\textbf{Algorithm} & \multicolumn{1}{l}{\textbf{\begin{tabular}[c]{@{}l@{}}Feature \\ Extractor\end{tabular}}} & \textbf{Input Size} & \textbf{{AP}} & \textbf{TP}   & \textbf{FN}   & \textbf{FP} & \textbf{Precision} & \textbf{Recall} & \multicolumn{1}{l}{\textbf{\begin{tabular}[c]{@{}l@{}}F1 \\ Score\end{tabular}}} & \textbf{FPS} & \multicolumn{1}{c}{\textbf{\begin{tabular}[c]{@{}c@{}}{Inference} \\ {Time (ms)}\end{tabular}}}\\ 
Faster R-CNN        & Inception v2               & 600 $\times$ 816 (variable)            & 0.202         & 1780          & 6351          & 1813        & 0.495              & 0.219           & 0.304             & 19.2 & {52}\\ 
Faster R-CNN        & Inception v2               & 608 $\times$ 608 (fixed)            & \textbf{{0.317}}        & \textbf{{2916}}          & \textbf{{5215}}          & 2654        & 0.524              & \textbf{{0.359}}           & \textbf{{0.426}}    & 21.1  &   {47}    \\ 
Faster R-CNN        & Resnet50                   & 600 $\times$ 816 (variable)            & 0.219         & 1909 & 6222 & 2117        & 0.474              & 0.235  & 0.314   & 8.6   & {116}       \\ 
Faster R-CNN        & Resnet50                   & 608 $\times$ 608 (fixed)            & 0.123         & 2061 & 6070 & 2456        & 0.456              & 0.253  & 0.326   & 8.2    & {122}      \\ 
YOLOv3             & Darknet-53                 & 320 $\times$ 320 (fixed)            & {0.107}         & 876           & 7255          & 4           & 0.995              & 0.108           & 0.194   & \textbf{{23.3}}  & {\textbf{43}}        \\
YOLOv3             & Darknet-53                 & 416 $\times$ 416 (fixed)             & {0.195}         & 1583          & 6548          & \textbf{1}  & \textbf{0.999}     & 0.195           & 0.326  & 18.6  & {54}\\ 
YOLOv3             & Darknet-53                 & 608 $\times$ 608 (fixed)             & {0.194}         & 1581          & 6550          & 10          & 0.994              & 0.194           & 0.325  & 11.8          & {85} \\ 
\multicolumn{1}{l}{{YOLOv4}} & \multicolumn{1}{l}{{CSPDarknet-53}} & \multicolumn{1}{l}{{320 $\times$ 320 (fixed)}} & \multicolumn{1}{l}{{0.157}} & \multicolumn{1}{l}{{1278}} & \multicolumn{1}{l}{{6853}} & \multicolumn{1}{l}{{5}}  & \multicolumn{1}{l}{{0.996}} & \multicolumn{1}{l}{{0.157}} & \multicolumn{1}{l}{{0.272}} & \multicolumn{1}{l}{{21.1}} & {47}\\ 
\multicolumn{1}{l}{{YOLOv4}} & \multicolumn{1}{l}{{CSPDarknet-53}} & \multicolumn{1}{l}{{416 $\times$ 416 (fixed)}} & \multicolumn{1}{l}{{0.202}} & \multicolumn{1}{l}{{1646}} & \multicolumn{1}{l}{{6485}} & \multicolumn{1}{l}{{\textbf{1}}}  & \multicolumn{1}{l}{{\textbf{0.999}}} & \multicolumn{1}{l}{{0.202}} & \multicolumn{1}{l}{{0.337}} & \multicolumn{1}{l}{{18.5}} & {54}\\ 
\multicolumn{1}{l}{{YOLOv4}} & \multicolumn{1}{l}{{CSPDarknet-53}} & \multicolumn{1}{l}{{608 $\times$ 608 (fixed)}} & \multicolumn{1}{l}{{0.209}} & \multicolumn{1}{l}{{1701}} & \multicolumn{1}{l}{{6430}} & \multicolumn{1}{l}{{64}} & \multicolumn{1}{l}{{0.964}} & \multicolumn{1}{l}{{0.209}} & \multicolumn{1}{l}{{0.344}} & \multicolumn{1}{l}{{12.5}} & {80}\\ \bottomrule
\end{tabular}}

\end{specialtable}

\begin{paracol}{2}
\switchcolumn
\vspace{-6pt}

\subsubsection{{{Effect of the Learning Rate}}}
{In order to measure the sensitivity of each algorithm to the learning rate hyperparameter, we conducted additional experiments with different values of learning rates (10\textsuperscript{$-$5}, 10\textsuperscript{$-$4}, 10\textsuperscript{$-$3}, and 10\textsuperscript{$-$2}) for each of the main algorithms (Faster R-CNN with Inception-v2, Faster R-CNN with Resnet 50, and YOLOv3 {and YOLOv4 both} with input size 416 $\times$ 416), on each of the two datasets. Figure~\ref{fig:AP_lr} shows a high sensitivity of the AP (measured on the validation dataset) to the learning rate value chosen during training, {except for YOLOv4, which benefits from the cosine annealing scheduler described in Section~\ref{section-yolo-v4}}. A learning rate of 10\textsuperscript{$-$3} yields the best performance in most cases, except that, {on the Stanford dataset, Faster R-CNN, with Inception-v2,  and YOLOv4  show better results at lower learning rates (10\textsuperscript{$-$4} and 10\textsuperscript{$-$5} respectively)}. A learning rate of 10\textsuperscript{$-$2} gives poor results in all cases except {for YOLOv4 on both datasets, and }for Resnet50 on the PSU dataset. A learning rate of 10\textsuperscript{$-$1} was also   tested, but it led to a divergent loss. These results highlight the importance of trying different values of learning rates when comparing the performance of object detection algorithms. The results shown in Figure~\ref{fig:AP_lr} confirm the better performance of {YOLOv4/}YOLOv3 and Faster R-CNN, respectively, on the PSU and Stanford datasets, when the learning rate is well chosen.}

\subsubsection{{{Effect of the Anchor Scales}}}
{The anchor scales used for the two algorithms are the default values specified in \mbox{Table~\ref{tab:comparison-YOLO-FRCNN}}.~We suspected that the anchor values could be the reason for the poor performance of {the tested algorithms} on the Stanford dataset, so we subsequently conducted {four} additional experiments with a different set of anchor scales. For YOLOv3 {and YOLOv4,} the new anchor scales were calculated using K-means clustering on the Stanford training dataset, and yielded smaller anchor sizes (10 $\times$ 27, 25 $\times$ 16, 17 $\times$ 26, 18 $\times$ 35, 22 $\times$ 31, \mbox{35 $\times$ 23}, 23 $\times$ 38, 27 $\times$ 34, and 31 $\times$ 42).  For Faster R-CNN, we used anchor scales reduced by half (64 $\times$ 64, 128 $\times$ 128, and \mbox{256 $\times$ 256}, instead of the default 128 $\times$ 128, \mbox{256 $\times$ 256}, and 51 $\times$ 512). Table~\ref{tab:reduced_anchors} shows the results obtained after using these anchors, compared to the previous results obtained with the default anchors. The performance was markedly lower for YOLOv3 {(and to a much lesser extent YOLOv4)}, which indicates that  the YOLOv3 algorithm is very sensitive to the change of anchor scales{, whereas this sensitivity was mitigated in YOLOv4.} As for Faster R-CNN with Resnet50 as a feature extractor, the AP was slightly lower (20.7\% down from 21.9\%), while the average IoU dropped noticeably (25\% down from 47.7\%).
In contrast, Faster R-CNN with Inception-v2 as feature extractor was the only algorithm that showed better results with the reduced anchor scales. The two rightmost columns in Table~\ref{tab:reduced_anchors} show the average width and height of the predicted bounding boxes. We notice that the dependency between the anchor scales and the predicted sizes is not straightforward. The average predicted sizes are more affected by the size of ground-truth bounding boxes in the training dataset (72 $\times$ 152 in average, as shown in Table~\ref{tab:average_car_size})  and adapt poorly to the different ground-truth car sizes and aspect ratios in the testing dataset (60 $\times$ 90 in average), which explains the low performance of all the tested algorithms on the Stanford dataset specifically. 
Moreover, we can observe that, despite the fact that the default anchor scales for Faster R-CNN are overall larger than those of YOLOv3 {and YOLOv4}, the first algorithm yields the best AP values on the Stanford dataset, which indicates that smaller anchor scales are not the solution for the poor performance obtained on the Stanford dataset.}

\begin{figure}[H]
\begin{center}  
\includegraphics[width=11cm]{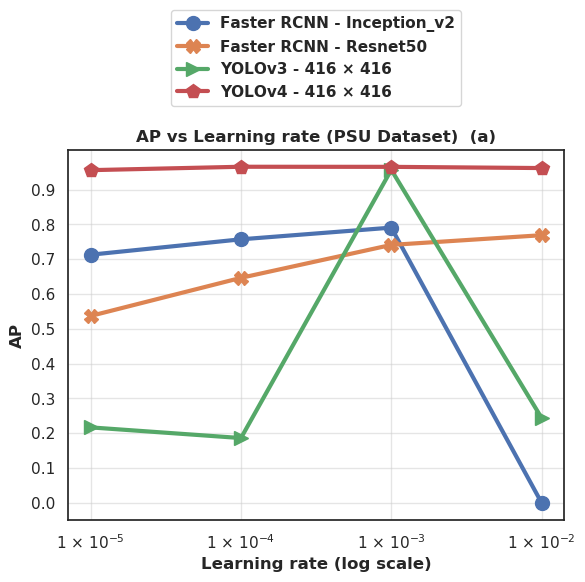}
\includegraphics[width=11cm]{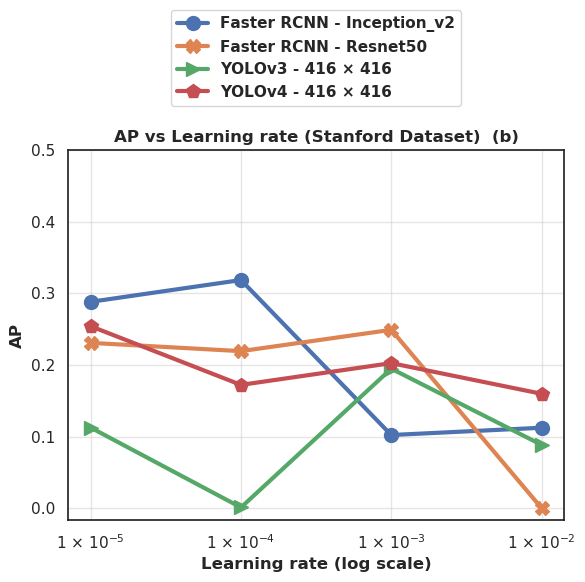}
\caption{Dependency  
between the AP (Average Precision) 
 and the learning rate, on the PSU 
 (a) and Stanford (b) datasets.
\label{fig:AP_lr}}  
\end{center}  
\end{figure}

\clearpage
\end{paracol}
\nointerlineskip
\begin{specialtable}[H]
\fontsize{8pt}{8pt}\selectfont 
\widetable
\caption{{Effect of reducing the anchor scales of {YOLOv4,} YOLOv3, and Faster R-CNN 
  on the Stanford Dataset.}} 
\label{tab:reduced_anchors}

\setlength{\cellWidtha}{\columnwidth/6-2\tabcolsep+0.0in}
\setlength{\cellWidthb}{\columnwidth/6-2\tabcolsep+1.4in}
\setlength{\cellWidthc}{\columnwidth/6-2\tabcolsep-0.3in}
\setlength{\cellWidthd}{\columnwidth/6-2\tabcolsep-0.4in}
\setlength{\cellWidthe}{\columnwidth/6-2\tabcolsep-0.4in}
\setlength{\cellWidthf}{\columnwidth/6-2\tabcolsep-0.3in}
\scalebox{1}[1]{\begin{tabularx}{\columnwidth}{>{\PreserveBackslash\raggedright\arraybackslash}m{\cellWidtha}>{\PreserveBackslash\raggedright\arraybackslash}m{\cellWidthb}>{\PreserveBackslash\raggedright\arraybackslash}m{\cellWidthc}>{\PreserveBackslash\raggedright\arraybackslash}m{\cellWidthd}>{\PreserveBackslash\raggedright\arraybackslash}m{\cellWidthe}>{\PreserveBackslash\raggedright\arraybackslash}m{\cellWidthf}}
\toprule                                                                              \textbf{Algorithm}                                                                                     & \textbf{Anchor Scales}                                                                                         & \textbf{\begin{tabular}[c]{@{}l@{}}AP \\(Average \\Precision)\end{tabular}} & \textbf{\begin{tabular}[c]{@{}l@{}}IoU \\(Intersection\\over Union)\end{tabular}} & \multicolumn{1}{l}{\textbf{\begin{tabular}[c]{@{}l@{}}Average \\ Predicted\\ Width\end{tabular}}} & \multicolumn{1}{l}{\textbf{\begin{tabular}[c]{@{}l@{}}Average \\ Predicted\\ Height\end{tabular}}} \\ \hline

\textbf{\begin{tabular}[c]{@{}l@{}}YOLOv3 416 $\times$ 416\\ (default anchors)\end{tabular}}                    & \begin{tabular}[c]{@{}l@{}}10 $\times$ 13, 16 $\times$ 30, 33 $\times$ 23, 30 $\times$ 61, 62 $\times$ 45, \\ 59 $\times$ 119, 116 $\times$ 90, 156 $\times$ 198, 373 $\times$ 326\end{tabular} & 0.195       & 0.89         & 67                                                                                                 & 170                                                                                                 \\ \hline
\textbf{\begin{tabular}[c]{@{}l@{}}YOLOv3 416 $\times$ 416\\ (reduced anchors)\end{tabular}}                    & \begin{tabular}[c]{@{}l@{}}10 $\times$ 27, 25 $\times$ 16, 17 $\times$ 26, 18 $\times$ 35, 22 $\times$ 31,\\ 35 $\times$ 23, 23 $\times$ 38, 27 $\times$ 34, 31 $\times$ 42\end{tabular}        & 0.082       & 0.55         & 127                                                                                                & 282                                                                                                 \\ \hline

\textbf{\begin{tabular}[c]{@{}l@{}}{YOLOv4 416 $\times$ 416}\\ {(default anchors)}\end{tabular}}                    & \begin{tabular}[c]{@{}l@{}}{12 $\times$ 16, 19 $\times$ 36, 40 $\times$ 28, 36 $\times$ 75, 76 $\times$ 55,} \\ {72 $\times$ 146, 142 $\times$ 110, 192 $\times$ 243, 459 $\times$ 401}\end{tabular} & {0.202}       & {0.92}         & {86}                                                                                                 & {170}                                                                                                 \\ \hline
\textbf{\begin{tabular}[c]{@{}l@{}}{YOLOv4 416 $\times$ 416}\\ {(reduced anchors)}\end{tabular}}                    & \begin{tabular}[c]{@{}l@{}}{10 $\times$ 27, 25 $\times$ 16, 17 $\times$ 26, 18 $\times$ 35, 22 $\times$ 31,}\\ {35 $\times$ 23, 23 $\times$ 38, 27 $\times$ 34, 31 $\times$ 42}\end{tabular}        & {0.188}       & {0.87}         & {81}                                                                                                & {192}                                                                                                 \\ \hline

\textbf{\begin{tabular}[c]{@{}l@{}}Faster R-CNN,\\ with ResNet50\\ (default anchors)\end{tabular}}     & \begin{tabular}[c]{@{}l@{}}Scales: 128 $\times$ 128, 256 $\times$ 256, 512 $\times$ 512\\ Aspect ratios: 1:1, 1:2, 2:1\end{tabular}       & 0.219       & 0.48         & 91                                                                                                 & 171                                                                                                 \\ \hline
\textbf{\begin{tabular}[c]{@{}l@{}}Faster R-CNN,\\ with ResNet50\\ (reduced anchors)\end{tabular}}     & \begin{tabular}[c]{@{}l@{}}Scales: 64 $\times$ 64, 128 $\times$ 128, 256 $\times$ 256\\ Aspect ratios: 1:1, 1:2, 2:1\end{tabular}         & 0.207       & 0.25         & 72                                                                                                 & 131                                                                                                 \\ \hline

\textbf{\begin{tabular}[c]{@{}l@{}}Faster R-CNN,\\ with Inception-v2\\ (default anchors)\end{tabular}} & \begin{tabular}[c]{@{}l@{}}Scales: 128 $\times$ 128, 256 $\times$ 256, 512 $\times$ 512\\ Aspect ratios: 1:1, 1:2, 2:1\end{tabular}       & 0.202       & 0.48         & 74                                                                                                 & 140                                                                                                 \\ \hline
\textbf{\begin{tabular}[c]{@{}l@{}}Faster R-CNN,\\ with Inception-v2\\ (reduced anchors)\end{tabular}} & \begin{tabular}[c]{@{}l@{}}Scales: 64 $\times$ 64, 128 $\times$ 128, 256 $\times$ 256\\ Aspect ratios: 1:1, 1:2, 2:1\end{tabular}         & 0.255       & 0.50         & 92                                                                                                 & 174                                                                                          \\ \bottomrule
\end{tabularx}}
\end{specialtable}
\begin{paracol}{2}
\switchcolumn
\vspace{-9pt}

\subsubsection{{Main Lessons Learned}}
Tables \ref{tab:Detailed_results_PSU} {and \ref{tab:Detailed_results_Stanford}} present the detailed results of all tested configurations of the two algorithms {on the PSU and Stanford datasets respectively}. The best performance for each metric, and each dataset is highlighted {in bold}. {We notice that {YOLOv4 with a medium input size (416 $\times$ 416)} and Faster R-CNN (with Inception-v2 feature extractor and a fixed input size) show the best results in terms of AP and recall, on the PSU and Stanford datasets, respectively.   In terms of precision, Faster R-CNN (with Resnet50 feature extractor and a variable input size) and YOLOv3{/YOLOv4} with a medium input size (416 $\times$ 416) perform better on the PSU and Stanford datasets, respectively.}
Figures \ref{fig:AP_time_PSU} and  \ref{fig:AP_time_Stanford} summarize the main results of this comparison study. They compare  the trade-off between AP and inference time for YOLOv3{/YOLOv4} (with 3 different input sizes) and Faster R-CNN (with two different feature extractors)  on the PSU and Stanford datasets, respectively{, with the default hyperparameters specified in Section~\ref{hyperparams}}. It can be observed that, while Faster R-CNN (with Inception v2 as feature extractor) gave the best trade-off in terms of AP and inference speed on the Stanford dataset {(followed closely by YOLOv4 416 $\times$ 416), YOLOv4} (with input size 320 $\times$ 320) presented the best trade-off on the PSU dataset. This lays emphasis on the fact that none of these algorithms outperforms the others in all cases, and that the best trade-off between AP and inference time depends on the characteristics of the dataset (object size, resolution, quality of annotation, {representativity of the training dataset,} etc.).

{In addition, while YOLOv4 has shown a steep increase in AP on the COCO dataset (from 33\% to 43\%), no such gap has been observed in our experiments on the smaller PSU and Stanford datasets, which indicates that the new features introduced in YOLOv4 were mainly tailored for the COCO dataset and may not be equally beneficial on other datasets.}

Finally, it should be noted that, although the present case study was restricted to only car objects, its conclusions can be easily generalized to similar types of objects in aerial images, since we did not use any specific feature of cars.

\section{Conclusions} \label{sec5}
In this study, we conducted a thorough experimental comparison of the {three} leading object detection algorithms ({YOLOv4,} YOLOv3, and Faster R-CNN) on two UAV imaging datasets that present very different characteristics, which makes the comparison more robust. Furthermore, the performance of the {three} algorithms was assessed using several metrics (mAP, IoU, FPS, AR\textsuperscript{max=1}, AR\textsuperscript{max=10}, and AR\textsuperscript{max=100},...) in order to uncover their strengths and weaknesses. One of the main conclusions that we can draw from this comparative study is that the performance of these algorithms largely depends on the characteristics of the dataset {and the representativity of the training images}. In fact, while Faster R-CNN (with Inception v2 as feature extractor) gave the best trade-off in terms of AP {(52\% higher than YOLOv4)} and inference speed {(only 10\% slower than YOLOv4)} on the Stanford dataset, {YOLOv4} (with an input size of 320 $\times$ 320) presented the best trade-off on the PSU dataset {(31\% more accurate and 2.4 times faster than Faster R-CNN)}. {The two tested feature extractors for Faster R-CNN yielded close results in terms of accuracy, while Inception v2 was 1.5 to 2.6 times faster than Resnet50. On the other hand,} the difference in accuracy between YOLOv3 and YOLOv4 was shown to be statistically insignificant on the Stanford and PSU datasets{, while they both show a high dependency to the input size (up to 1.9 times slower when passing from 320 $\times$ 320 to 608 $\times$ 608). In addition, we have shown that a badly chosen learning rate can yield extremely low AP (almost 0), and that the choice of the anchor scale values can impact the AP up to 58\% for YOLOv3, and 26\% for Faster R-CNN.} 
{As future work, we intend to extend our results to the newly released EfficienDet~\cite{efficientdet} detector  and to much larger datasets of aerial images}.

\vspace{6pt} 



\authorcontributions{Conceptualization, A.K. and A.A.; methodology, A.A., B.B., and A.K.; software, M.A., A.S., A.A., and B.B.; validation, A.A. and A.K.; formal analysis, A.A., B.B.,  and A.K.; investigation, A.A., A.K., M.A.,  and A.S.; resources, A.K., A.A., M.A., and A.S.; data curation, M.A. and A.S.; writing--original draft preparation, A.A.; writing---review and editing, A.A. and B.B.; visualization, A.A., M.A., and A.S.; supervision, A.K. and A.A.; project administration, A.K.; funding acquisition, A.K. All authors have read and agreed to the published version of the manuscript.}

\funding{\textls[-15]{This work is supported by the research grant SEED-2020-05 from Prince Sultan University. }}



\dataavailability{The PSU dataset used in this study is available at:

https://github.com/aniskoubaa/psu-car-dataset.} 

\acknowledgments{\textls[-15]{The authors would like to acknowledge the support of Prince Sultan University for paying the Article Processing Charges (APC) of this publication.
 We also thank Taha Khursheed for working on the prior conference version of this paper.}}

\conflictsofinterest{The authors declare no conflict of interest. The funders had no role in the design of the study; in the collection, analyses, or interpretation of data; in the writing of the manuscript; or in the decision to publish the results.}


\end{paracol}
\reftitle{References}

\end{document}